\pdfoutput=1

\documentclass[11pt]{article}

\usepackage[final]{acl}

\usepackage{times}
\usepackage{latexsym}

\usepackage[T1]{fontenc}

\usepackage[utf8]{inputenc}

\usepackage{microtype}

\usepackage{inconsolata}

\usepackage{times}
\usepackage{latexsym}
\usepackage[T1]{fontenc}
\usepackage[utf8]{inputenc}
\usepackage{microtype}
\usepackage{inconsolata}
\usepackage{graphicx}

\usepackage{tipa}
\usepackage{makecell}
\usepackage{rotating}
\usepackage{amssymb}
\usepackage{pifont}
\usepackage{tabularray}
\usepackage{array}
\usepackage{fdsymbol}
\usepackage{cleveref}
\usepackage{multirow}
\usepackage[scaled=0.85]{helvet}
\usepackage{tcolorbox}
\usepackage{xspace}
\usepackage{booktabs}
\usepackage{tabularx}
\usepackage{caption}
\usepackage{subcaption}
\usepackage{setspace}
\usepackage{cellspace}
\usepackage{twemojis}
\usepackage{tikz}

\renewcommand{\arraystretch}{1.05}

\newcommand{\cellcomp}[2]{%
  \pgfmathparse{#1-#2}%
  \ifdim\pgfmathresult pt>0pt
    \cellcolor{green!18}{#1}%
  \else
    \cellcolor{red!15}{#1}%
  \fi
}

\newcolumntype{Y}{>{\raggedright\arraybackslash}X}

\makeatletter
\DeclareRobustCommand*{\escapeus}[1]{%
  \begingroup\@activeus\scantokens{#1\endinput}\endgroup}
\begingroup\lccode`\~=`\_\relax
\lowercase{\endgroup\def\@activeus{\catcode`\_=\active \let~\_}}
\makeatother
\newcommand{\myemph}[1]{\textsf{{\escapeus{#1}}}}

\definecolor{cambridgeblue}{RGB}{163,193,173}
\definecolor{customyellow}{HTML}{E0A000} 
\definecolor{cambridgewarmblue}{RGB}{0,189,182}

\usepackage{tabularray}
\usepackage{cellspace}
\newcommand\cincludegraphics[2][]{\raisebox{-0.3\height}{\includegraphics[#1]{#2}}}
\usepackage{setspace}
\usepackage{twemojis}

\newtcbox{\characterhighlight}{on line, colback=cambridgeblue!50, boxrule=0.2mm, left=0.5mm, right=0.2mm, top=0.2mm, bottom=0.2mm}

\newtcbox{\spacehighlight}{on line, colback=customyellow!50, boxrule=0.2mm, left=0.5mm, right=0.2mm, top=0.2mm, bottom=0.2mm}

\newtcbox{\phonemehighlight}{on line, colback=cambridgewarmblue!50, boxrule=0.2mm, left=0.5mm, right=0.2mm, top=0.2mm, bottom=0.2mm}

\definecolor{customgreen}{HTML}{3BAE2E}  
\definecolor{customred}{HTML}{D13438}  
\definecolor{customyellow}{HTML}{E0A000} 

\title{Looking to Learn: Token-wise Dynamic Gating for Low-Resource Vision-Language Modelling}

\author{
    {\bf Bianca-Mihaela Ganescu}\thanks{   \textbf{ Corresponding Authors:} \texttt{bmg44@cam.ac.uk, sas245@cam.ac.uk}}~\texttwemoji{baby_bottle}\texttwemoji{mag} \quad
    {\bf Suchir Salhan*}~\texttwemoji{baby_bottle}\texttwemoji{mag} \quad
    {\bf Andrew Caines}~\texttwemoji{baby_bottle}\texttwemoji{mag} \quad
    {\bf Paula Buttery}~ \texttwemoji{baby_bottle}\texttwemoji{mag} \\
\texttwemoji{baby_bottle}  ALTA Institute \texttwemoji{mag} Department of Computer Science \& Technology, University of Cambridge
}

\begin{document}
\maketitle
\begin{abstract}
Training vision-language models on cognitively-plausible amounts of data requires rethinking how models integrate multimodal information. Within the constraints of the Vision track for the BabyLM Challenge 2025, we propose a lightweight decoder-based architecture with (1) token-wise dynamic gating for adaptive fusion of linguistic and visual cues, (2) feature modulation and channel attention to maximise the utility of limited visual information and (3) auxiliary contrastive objectives for visual grounding. 
Evaluation on five benchmarks (BLiMP, BLiMP Supplement, EWoK, Winoground and VQA) shows competitive or superior performance to multimodal baselines. More notably, our dynamic gate discovers interpretable patterns without explicit supervision, favouring visual cues for content words and linguistic cues for function words. While we identify limitations in the Challenge constraints, such as the information bottleneck created by global image embeddings and training instability from the dataset split, our findings establish dynamic gating as a powerful tool for efficient multimodal learning, offering both interpretability and performance even under severe constraints.
\end{abstract}

\noindent
\begin{tblr}{colspec = {Q[c,m]|X[l,m]}, stretch = 0}
    \cincludegraphics[width=1.35em, keepaspectratio]{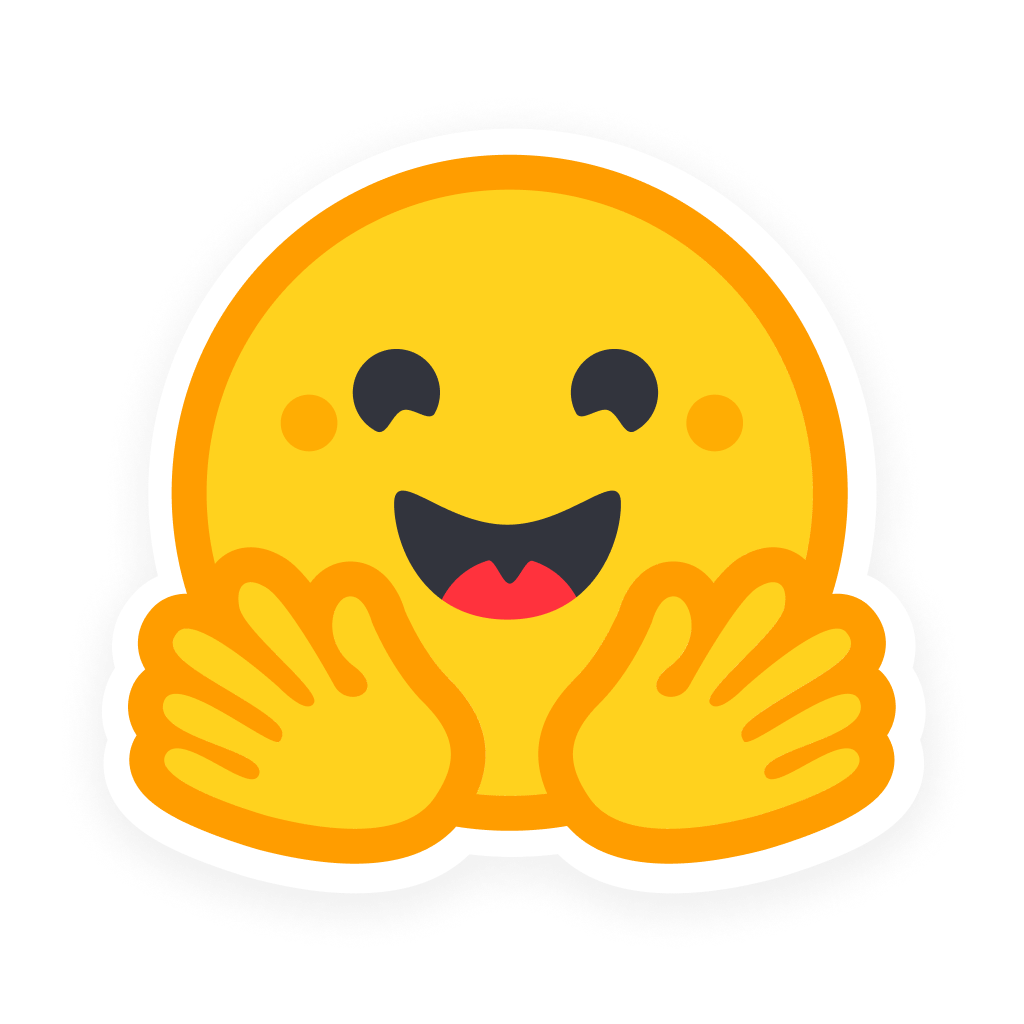} &
    \setstretch{.5}{\small\myemph{\textbf{LookingtoLearn} on \href{https://huggingface.co/LookingtoLearn}{HuggingFace} (models, tokenizers, and checkpoints)}} \\

    \cincludegraphics[width=1.1em, keepaspectratio]{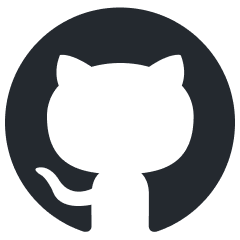} &
    \setstretch{.5}{\small\myemph{Training Code Open-Sourced on \href{https://github.com/biancaganescu/looking-to-learn}{GitHub}}}
\end{tblr}

\section{Introduction}

Large language models have achieved impressive capabilities, yet their learning process markedly differs from naturalistic human language learning. Children learn their first language from just tens of millions of words \citep{babylm_call_for_papers_1, gilkerson2017mapping} with minimal supervision, whereas state-of-the-art language models require three to four magnitudes more data \citep{babylm_call_for_papers_1}.

\begin{figure}[t!]
    \centering
    \includegraphics[width=\linewidth]{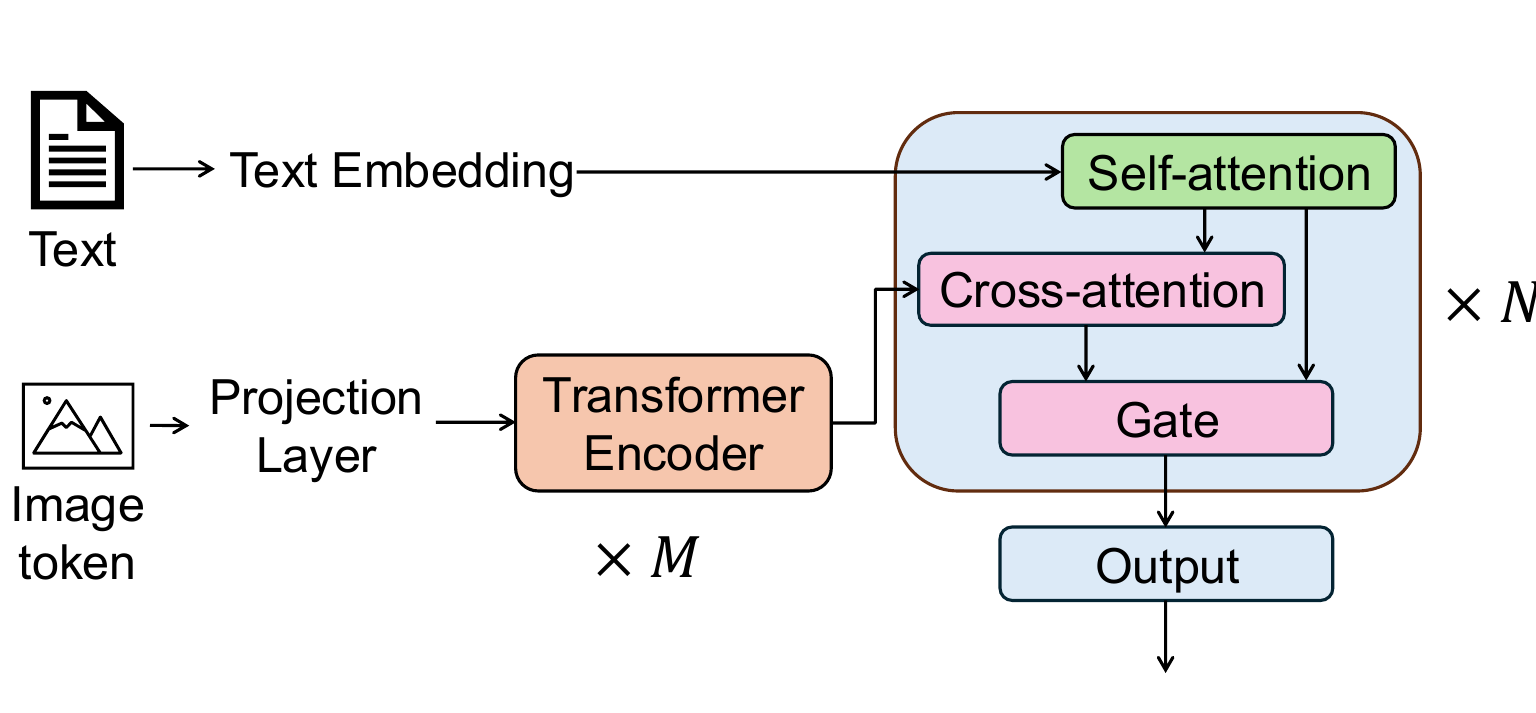}
    \caption{Simplified dual-stream architecture. The \textbf{text processing stream} (top) embeds text input tokens and feeds them through an $N$-layer transformer decoder, which applies masked self-attention, cross-attention to image features and a dynamic gating module to fuse representations. The \textbf{image processing stream} (bottom) projects a DINOv2 global token into the same space and processes it with an $M$-layer transformer encoder. The image processing stream, cross-attention and gating modules are skipped for text-only samples.}

    \label{fig:architecture_overview}
\end{figure}

Furthermore, human language learning is inherently a multimodal process. Usually, visual experiences play a crucial role in the acquisition of early language and its expansion in the first years of life \citep{rose2009cognitive, morgenstern2014children, morgenstern2023children, karadoller2024first}. This cognitive reality motivates our work in the BabyLM Challenge Vision track \citep{babylm_call_for_papers_3}, where we develop a framework inspired by human selective attention that learns \textit{when} and \textit{how} to leverage visual cues during language processing without explicit supervision.

Our proposed solution is a decoder-based vision-language model for which we introduce three key innovations. First, we implement a \textbf{dynamic gating mechanism} that learns to selectively weight visual versus linguistic cues for each token based on context. Second, we explore several \textbf{feature enhancement techniques} in order to maximise the utility of limited visual information. Third, we investigate the impact of \textbf{contrastive learning auxiliary objective functions} that operate at both the sentence and word levels under low-resource constraints.

In this work, we aim to answer several key questions: 
\begin{enumerate}
    \item Q: Can dynamic gating mechanisms be repurposed to learn meaningful vision-language fusion patterns without explicit supervision? (Subsection \ref{dynamic_gating}) \\
    A: Yes, statistical analysis of our dynamic gate's outputs shows a strong correlation between branch selection and parts-of-speech, as well as a weak correlation between branch selection and concreteness and imageability scores. (Section \ref{interpretability})

     \item Q: If so, which linguistic phenomena do our models prioritise visual information for, and does this align with human word grounding? \\
     A: We find that for parts-of-speech which are open-class and tend to be more grounded (adjective, noun, proper noun, verb) \citep{haley2024grounded}, the dynamic gate assigns more weight to visual signals than for function words (conjunction, punctuation, symbols, auxiliary verbs, particles) which tend to be less grounded \citep{haley2024grounded}. (Section \ref{interpretability})

    \item Q: Is the setup of the Vision track optimal for multimodal learning? In particular, can architectural mechanisms compensate for the limited visual information provided by global image embeddings? (Subsection \ref{feature_representation}) \\
    A: In our framework, global image embeddings create an information bottleneck that feature enhancements cannot fully address (Subsection \ref{res:feature_representation}). Moreover, we find that the split between text-only and image-caption data causes training instability and identify misalignments between the training data and multiple evaluation benchmarks. (Section \ref{discussion})

    \item Q: Do contrastive learning auxiliary objectives help or hinder small vision-language models under significant data constraints? (Subsection \ref{objectives}) \\
    A: Contrastive learning objectives prove counterproductive without sufficient scale for our selected benchmarks. (Subsection \ref{res:objective_functions})

\end{enumerate}

Performance analysis on five BabyLM Challenge benchmarks (BLiMP \citep{warstadt2020blimp}, BLiMP Supplement \citep{warstadt2025findings}, EWoK \citep{ivanova2024elements}, Winoground \citep{thrush2022winoground} and VQA \citep{goyal2017making}) reveals task-specific benefits of our proposed framework, with our base model achieving competitive or superior performance compared to the multimodal baselines of the 2025 Challenge.

Overall, this work contributes to the broader goal of taking inspiration from human learning for the development of language models, not just in terms of data, but also in their underlying mechanisms. While the results of dynamic gating show that architectural design can lead to meaningful patterns, our findings also reveal which constraints (visual representation, data curriculum, training datasets and evaluation benchmarks) must be addressed next in order to further improve vision-language models based on human learning.

\section{Background}

\subsection{Vision-Language Models}

\textbf{Vision-language models (VLMs)} combine an image encoder and (optionally) a text encoder with a multimodal fusion module to learn joint representations for tasks such as captioning, retrieval, and visual question-answering. Although specific VLM architectures vary, most share a \textbf{vision encoder} projecting images into embedding features aligned with language model embeddings, often implemented as a Vision Transformer (ViT) patch encoder pre-trained on rich visual datasets \citep{dosovitskiy2021imageworth16x16words}, and a \textbf{text encoder}. While early VLMs (e.g.,  CLIP \citep{radford2021learning} and ALIGN \citep{jia2021scaling}) use both vision and text encoders trained jointly using contrastive learning to align visual and textual representation in a shared latent space \citep{li2025benchmark}, more recent models such as LLaVA \citep{liu2023visual} no longer employ a separate text encoder, but simply use a visual encoder with a text decoder. 

\subsubsection{GIT and Flamingo Baselines}

The baselines used in the Vision track are the Flamingo \citep{alayrac2022flamingo} and the Generative Image-to-text Transformer (GIT) \citep{wang2022git} vision-language models. 

 The GIT \citep{wang2022git} architecture consists of an image encoder and a text decoder. The image encoder is pre-trained using a contrastive learning objective, and outputs visual features that are linearly projected and concatenated with embedded text tokens to form the input to the decoder. The entire model is trained using next token prediction, where each token is predicted based on both the preceding text tokens and the visual features. 
 
Flamingo \citep{alayrac2022flamingo} is a decoder-based multimodal model that interleaves text decoder layers with gated cross-attention dense blocks that incorporate visual input. An image encoder extracts visual features, which a Perceiver  Resampler module \citep{jaegle2021perceiver}
compresses into a fixed number of tokens per image. These serve as keys and queries in the gated cross-attention dense layers inserted between language model blocks, where a tahn-gated learnable scalar scales each cross-attention and feed-forward sublayer to control the flow of visual information. The model is trained with next token prediction. 

Last year's submissions to the BabyLM Challenge Vision track did not beat the Flamingo and GIT baselines \citep{babylm_proceedings_2}. \citet{alkhamissi-etal-2024-dreaming} proposed a self-synthesis strategy for training a BabyLLaMA \citep{timiryasov2023baby} model, comprising four phases ranging from basic language skills to cognitive tasks. \citet{saha-etal-2024-exploring} evaluated the effect of curriculum learning on GIT and Flamingo, concluding that benefits were architecture-, training- and task-dependent. \citet{ klerings-etal-2024-developmentally} investigated the role of visual data in language learning for the GIT architecture, finding that visual training data improves model performance on multimodal benchmarks but has no effect on text-only benchmarks. Moreover, they analysed task-specific neuron usage and concluded that their models are highly modular, with visual inputs influencing which components the model uses to process the same text input.  

We hypothesise that the GIT and Flamingo models, originally proposed for large-scale training, lack explicit cognitive motivation and may underperform when scaled down, prompting the implementation of our architecture. We point out that our dynamic gating approach contrasts with Flamingo's gated cross-attention dense blocks as follows: firstly, Flamingo applies uniform layer-wise gating parameters across all tokens, whereas our dynamic gate adapts based on individual tokens. Secondly, our model consistently injects visual features at every decoding layer, while Flamingo introduces visual information every few layers, which could limit the model’s ability to learn a stable textual representation.

It is worth noting that the 2024 Challenge did not impose a limit on the number of training epochs, and that the 2024 Flamingo and GIT baselines were trained on 20 epochs worth of text-only data and potentially up to 80 epochs worth of image-caption data. Similarly, the 2025 baselines were trained using a 1:4 ratio between text-only and image-caption data, while respecting the 10-epoch training limit.

\subsection{Token-Level Integration of Visual and Linguistic Information}

In this work, we ask whether dynamically weighting different modalities can be used to improve next token selection in autoregressive models. \citet{wang2018learning} and \citet{kiela-etal-2018-dynamic} asked a similar question about how to dynamically weight linguistic and visual input based on word type. However, their goal was to create static word embeddings relying on weak supervision. In contrast, we propose using dynamic gating as an unsupervised mechanism during autoregressive generation, where the model decides at each step which modality should produce the next token.

It has been shown in cognitive science research that concrete and abstract words are differentiated in human processing \citep{binder-2005}. \citet{wang2010neural} found that abstract words primarily activate language-related brain regions, whereas concrete words engage perceptual brain areas. Further work showed that functional Magnetic Resonance Imaging patterns for concrete nouns can be decoded by both linguistic and visual representations \citep{anderson2017visually}, but that abstract nouns are only decodable via linguistic representations \citep{wang2018learning}. 

An advantage has been reported for words judged to be more concrete than those judged to be more abstract in the learning of distributed semantic representations \citep{bruni-2014,hill-korhonen-2014-learning}. This means that language models are likely to perform better on benchmarks containing more concrete words \citep{pezzelle-etal-2021-word}. As \citet{kiela-etal-2018-dynamic} observe, there are complex interactions between concreteness and other factors such as word frequency and word class. We examine concreteness and word class separately in this work and leave the study of their interaction, along with word frequency, for future work.

\section{Method}

\subsection{Overview} 
We present a multimodal framework for the BabyLM Vision track that learns language from both ungrounded and visually-grounded text data. We assume that the training data and number of training epochs are fixed according to the \textbf{constraints of the BabyLM Challenge 2025}, while the architecture and training regime are variables which we aim to optimise. Specifically, we develop a \textbf{dual-stream transformer architecture with three key innovations} that aim to mirror human language processing and improve model performance under Challenge constraints:
\begin{enumerate}
    \item \textbf{Cognitive alignment through dynamic gating}: Unlike standard vision-language models that use uniform fusion strategies, we implement a token-wise dynamic gating mechanism with four variants exploring different granularities and decision levels. This mechanism learns to adaptively weight visual versus linguistic information for each token, drawing inspiration from how humans selectively integrate multimodal information.
    
    \item \textbf{Maximising limited visual information}: Given the constraint of using only a global image embedding during training, we implement multiple strategies to compensate for limited visual information. These include modulation techniques that dynamically transform features based on cross-modal context, and channel attention to identify salient aspects within the limited visual representation.
    
    \item \textbf{Visual grounding via auxiliary objective functions}: we explore two auxiliary objective functions to enhance visual grounding in our framework: (1) a contrastive learning objective \citep{radford2021learning} which aligns entire captions with images at the sentence level, and (2) LexiContrastive Grounding \citep{zhuang2024lexicon}, which performs word-level alignment between individual tokens and images. These auxiliary objectives aim to improve language learning by creating stronger associations between linguistic and visual representations. 

 \end{enumerate}

\subsection{Base Architecture} \label{base_architecture}

At the core of our framework, we design an autoregressive dual stream transformer drawing inspiration from the architecture of state-of-the-art vision-language models such as LLaVA \citep{liu2023visual} and QWen-VL \citep{bai2023qwenvlversatilevisionlanguagemodel}. A simplified illustration of our architecture is shown in Figure \ref{fig:architecture_overview}.

The architecture consists of four main components: (1) a \textbf{text processing stream} that uses standard decoder layers with learned embeddings and positional encodings to process both text-only data and image captions; (2) an \textbf{image processing stream} that takes DINOv2 embeddings as input, projects them  into the model’s hidden space and refines them with additional transformer encoder layers for empirical performance, future patch-token compatibility and computational efficiency (details in Appendix \ref{appendix:diff_encoders}); (3) a \textbf{multimodal decoder} integrates text and image features using three sequential mechanisms: masked self-attention applied to the text features, followed by cross-attention fusion between text and image features (when available), and dynamic gating that adaptively determines how much to rely on visual versus linguistic information for each token; (4)  an \textbf{output projection layer} that maps the decoder outputs back to the vocabulary space for next token prediction.

\subsection{Dynamic Gating} \label{dynamic_gating}
While dynamic gating in multimodal AI has primarily focused on classification tasks, demonstrating improved robustness and computational efficiency \citep{xue2023dynamic, wang2024dynamic, xie2020gate}, we ask whether this idea can be repurposed as a cognitively-motivated mechanism for token selection in multimodal autoregressive models. 

Our approach is hypothesis-driven. Dynamic gating has proven effective in other multimodal settings, and large vision-language models appear to exhibit \textit{implicit} gating capabilities through attention patterns learned at scale. We therefore ask: does introducing an \textit{explicit} token-wise gating mechanism help smaller, data-efficient models achieve similar selective integration more reliably?

Our hypothesis is therefore as follows: just as human language processing selectively integrates visual information, for example, relying heavily on visual inputs for concrete, perceptual words (e.g., ``dog", ``red") while defaulting to linguistic knowledge for abstract terms (e.g., ``therefore", ``impossible"), a token-wise dynamic gating mechanism could teach a model to make similar fine-grained fusion decisions.  By conditioning each gate on both the current text hidden state and the cross-attention features, the model can learn to amplify the vision input when it truly informs the next word and ignore it when it does not.

We implement four variants of a dynamic gating mechanism, varying along two axes: \textbf{(1) granularity}, whether the gate is computed per feature or per token, and \textbf{(2) soft vs hard}, whether the gate outputs continuous weights or discrete decisions. The granularity axis investigates whether different tokens require different subsets of visual features (e.g., colour features for ``red", spatial features for ``above") or whether coarse per-token gating is sufficient. The soft vs hard gating axis examines whether binary selection or continuous weighting of features yields more interpretable fusion patterns and better performance. 

More specifically, the four gating variants compute gating weights $g$ to dynamically fuse text and cross-attention representations via
\begin{equation}
  h_{\text{fused}}
  =  g \,\odot\, h_{\text{text}}
    + \bigl(1 -  g\bigr)\,\odot\, h_{\text{crossAttn}}
\end{equation}

\noindent
and differ in whether $g$ operates per feature or per token and makes continuous or discrete selections.
The technical implementation details for the four gating variants are available in Appendix \ref{ch4:dynamic_gating}.

\subsection{Feature Representation} \label{feature_representation}
The BabyLM Challenge provides the images in the training data as single global embeddings. While computationally efficient, this approach limits the spatial visual information available to the model. Traditional vision-language models benefit from patch token representations that preserve spatial information and enable fine-grained visual grounding \citep{dosovitskiy2021imageworth16x16words}. Therefore, the next aspects we investigate in this work are methods of maximising the utility of the global image tokens provided in the Challenge. 

We explore two complementary modulation techniques, FiLM \citep{perez2018film} and DyIntra \citep{gao2019dynamic}, which dynamically reshape one set of features based on another, as well as a global channel-attention enhancement. These approaches target different aspects of the representation bottleneck: modulation techniques address cross-modal feature interaction, while channel attention addresses intra-modal feature refinement. 

While the dynamic gating mechanism determines \textit{how much} information to incorporate from each modality, FiLM and DyIntra determine \textit{how} that information should be transformed, and a channel attention mechanism determines \textit{what} is meaningful within the image features. 

We evaluate these methods at several integration points within our architecture to determine which approach most effectively compensates for the lack of spatial visual information.

The technical details of our implementations are available in Appendix \ref{ch4:feature_representation}.

\subsection{Auxiliary Objective Functions} \label{objectives}
As previous work in vision-language models suggests \citep{lu202012}, a multi-task objective can improve model performance. In this work, we explore training our models using two auxiliary functions, Contrastive Language-Image Pre-training (CLIP) \citep{radford2021learning} and LexiContrastive Grounding (LCG) \citep{zhuang2024lexicon}. Both functions aim to ground textual representations in visual concepts through contrastive learning, creating a shared embedding space where semantically related image-text pairs are positioned closer together. However, they operate at different levels of granularity: CLIP aligns entire captions with their corresponding images at the sentence level, while LCG performs alignment at the word level between individual tokens and images.

 Recent research shows that visual grounding at both sentence and word levels can improve word acquisition in low-data regimes \citep{zhuang2023visual}. A CLIP objective could capture global associations that support contextual understanding, while an LCG objective might reflect fine-grained grounded learning. However, as we discuss in the results section (Section \ref{res:objective_functions}), the effectiveness of these auxiliary objectives significantly depends on various factors, including the visual representation format,  batch size and training data.

The technical details of each auxiliary objective function are available in Section \ref{ch4:objective_functions}.

\section{Results}
\begin{table*}[t]
  \footnotesize
  \centering
  \renewcommand{\arraystretch}{1}  
\begin{tabularx}{\textwidth}{
     p{4.9cm}*{5}{>{\centering\arraybackslash}p{1.882cm}}
  }
    \toprule
    \textbf{Model} &
    \textbf{BLiMP} &
    \textbf{BLiMP Supplement} &
    \textbf{EWoK} &
    \textbf{Winoground} &
    \textbf{VQA*} \\
    \midrule
    \multicolumn{6}{l}{\textbf{Baselines 2025}} \\
    Flamingo (BabyLM Challenge 2025) & \cellcolor{blue!13}{70.9} & \cellcolor{blue!13}{65.1} & \cellcolor{blue!13}{51} & \cellcolor{blue!13}{54.8} & \cellcolor{blue!13}{43.31} \\
    GIT (BabyLM Challenge 2025) & \cellcolor{blue!13}{72.2} & \cellcolor{blue!13}{66.4} & \cellcolor{blue!13}{51.8} & \cellcolor{blue!13}{56.2} & \cellcolor{blue!13}{49.82} \\
    \midrule
    \multicolumn{6}{l}{\textbf{Baselines 2024}} \\
    Flamingo (BabyLM Challenge 2024) &  70.9 & 65.0 & 52.7 & 51.6 & 52.3 \\
    GIT (BabyLM Challenge 2024) & 65.2 & 62.7 & 52.4 & 55.5 & 54.1 \\
    BabyLLaMA \citep{alkhamissi-etal-2024-dreaming} & 72.9 & 54.2 & 50.2 & 50.9 & 42.0\\
    $\text{Flamingo}_{\text{CL}}$ T+C \citep{saha-etal-2024-exploring} & 60.13 & 53.28 & 50.71 & 50.80 & 40.85\\
    $\text{GIT}_{\text{CL}}$ T+C \citep{saha-etal-2024-exploring} & 64.05 & 51.24 & 50.98 & 55.23 & 43.98\\
    GIT 1/0.25 \citep{klerings-etal-2024-developmentally} & 71.2 & 64.6 & 52.5 & 56.2 & 52.2\\
    GIT 1/0.125 \citep{klerings-etal-2024-developmentally} & 66.3 & 61.7 & 52.3 & 57.0 & 52.6\\
    \midrule
    \multicolumn{6}{l}{\textbf{Our framework}} \\
    Base, soft gate per feature  & \cellcomp{74.33}{0} & 56.36 & 50.81 & 51.61 & \cellcomp{50.02}{0} \\
    \midrule
    \multicolumn{6}{l}{\textbf{Architectural features}} \\
    Soft gate per token & \cellcomp{73.86}{0} & 55.43 & 51.56 & 52.14 & 48.39 \\
    Hard gate per feature &  \cellcomp{74.10}{0} & 54.16 & 51.20 & 50.13 & 45.62 \\
    Hard gate per token & \cellcomp{74.19}{0} & 54.59 & 51.16 & 50.80 &  45.51 \\
    No gate &  \cellcomp{74.70}{0} & 55.75 & 50.77 &  51.34 & \cellcomp{50.58}{0} \\
    FiLM on text & \cellcomp{74.32}{0} & 55.10 & 50.61 & 53.49 & 46.04 \\
    FiLM on cross-attention & \cellcomp{74.95}{0} & 56.36 & 51.62 & 52.68 & 49.66 \\
    FiLM on image & \cellcomp{73.80}{0} & 54.59 & 51.06 & 50.13 & 17.92 \\
    DyIntra on text & \cellcomp{74}{0} & 56.97 & 51.73 & 51.47 & 47.16 \\
    DyIntra on cross-attention & \cellcomp{73.68}{0} & 56.68 & 51.57 & 53.22 &  48.87 \\
    DyIntra on image & \cellcomp{74.69}{0} & 56.57 & 51.28 & 50.00 & 45.61 \\
    Channel attention & \cellcomp{74.24}{0} & 54.23 & 51.15 & 51.15 & 49.15 \\
    \midrule
    \multicolumn{6}{l}{\textbf{Auxiliary objective functions}} \\
    NTP + CLIP & 72.28 & 54.35 & 51.45 & 51.47  & 47.72 \\
    NTP + LCG & 70.27  & 56.91 & 49.74 & 50.00 & 36.62 \\
        \bottomrule
\end{tabularx}

  \setlength{\abovecaptionskip}{4pt}
\caption{Performance of our base model and variants on five BabyLM Challenge benchmarks. Scores for our models and 2025 baselines are computed using the 2025 evaluation pipeline (BLiMP, BLiMP-S, EWoK, Winoground) and 2024 pipeline (VQA). Green shading indicates performance above the 2025 baselines.
\vspace{-15pt}}
\label{tab:main_results}
\end{table*}

To evaluate our framework, we select five of the benchmarks proposed in the BabyLM Challenge: BLiMP \citep{warstadt2020blimp} and BLiMP Supplement for grammar, EWoK \citep{ivanova2024elements} for world knowledge, Winoground \citep{thrush2022winoground} for vision-linguistic compositional reasoning and VQA \citep{goyal2017making} for image-based question answering. For the first four we used the 2025 evaluation pipeline\footnote{\url{https://github.com/babylm/evaluation-pipeline-2025}}, while for VQA we used the 2024 repository\footnote{\url{https://github.com/babylm/evaluation-pipeline-2024}}. A detailed description of each benchmark is available in Appendix \ref{appendix:background_evaluation_pipeline}. 

For all architectural features and training strategies we define, we conduct experiments in the form of ablation studies in order to evaluate each potential improvement in isolation. A summary of all the experiments we define is available in Table \ref{tab:experiment_summary}. We train all our models in the same conditions (as described in Table \ref{tab:hyperparameters_training})  using the same model hyperparameters (summarised in Table \ref{tab:hyperparameters}), with the exception of the auxiliary objective function, for which we increase the batch size from 64 to 128 as a larger batch size is recommended for contrastive learning \citep{chen2020simple}.

Our key observations from the complete set of results (Table \ref{tab:main_results}) are summarised below.

\subsection{Baselines}
\textbf{Our framework achieves a higher score on BLiMP (almost 4\% higher than Flamingo and over 2\% higher than GIT) and competitive scores for EWoK, Winoground and VQA.} We suggest that our base model outperforms Flamingo and GIT on BLiMP due to architectural differences. These include the clear separation between the text and image streams and consistent fusion in our base model. In our proposed architecture, the first decoder layer's self‐attention module processes only textual input, whereas GIT concatenates the image token(s) with the text input before they are fed into the model, which could introduce noise when extracting linguistic signals. Our model consistently integrates visual features at each decoding layer, whereas Flamingo incorporates visual information only at intermittent layers, which could affect the model’s ability to learn a robust textual representation.

The lower performance of our model on BLiMP Supplement is due to differences in training data. As shown in Appendix \ref{training_dynamics}, the image–caption dataset supports this benchmark far better than the text-only dataset. This favours Flamingo and GIT, trained with a 1:4 text-only/image–caption ratio versus our 1:1 ratio. Winoground shows a similar trend, with baselines benefiting from more image–caption training epochs.

\subsection{Performance of Architectural Features} \label{res:feature_representation}
\textbf{The dynamic gating modules maintain the performance of our base model without gating on BLiMP and BLiMP Supplement, bring modest benefits for Winoground and show mixed results on VQA.} As shown in Table \ref{tab:main_results}, our dynamic gating modules do not have a significant effect on the model's performance on BLiMP and BLiMP Supplement, which is the desired outcome for the text-only benchmarks. There is very little variation in the EWoK scores across the models, which we attribute to a mismatch between training and evaluation data and further discuss in Section \ref{limitations}.

The \textit{soft gate} and \textit{hard gate per token} models outperform the \textit{no gate} model on Winoground. We hypothesise that the gating mechanism in these models produces slightly cleaner, more discriminative joint representations between images and text, which in turn yields a small but consistent improvement. 

We observe limited performance benefits of the gating modules on VQA, with the \textit{hard gate} models achieving a lower score ($\sim5\%$ lower) compared to the \textit{no gate} variant. We hypothesise that the \textit{hard gates} may have learned to allow for stronger image signals than is optimal for VQA, especially since the image-captioning training set contains few constructions similar to VQA (see Section \ref{limitations}).

\textbf{Modulation and channel attention achieve mixed results over the five benchmarks, underscoring that the global image embedding represents a performance bottleneck.} Across the seven variants we implement, no single feature representation technique uniformly improves all five benchmarks (Table \ref{tab:main_results}). FiLM applied to textual representation and cross-attention, along with DyIntra applied to cross-attention, shows modest improvements on Winoground (+1.88\%, +1.07\%, and +1.61\% respectively) by potentially creating more separable joint representations. 

With varied impact, all techniques decrease the performance of our base model on VQA. Nevertheless, the cross-attention modulation and our attention seem to preserve most of the linguistic and visual signals needed for VQA, while also bringing slight improvements on Winoground. The results collectively demonstrate that feature modulation and channel attention techniques designed for rich representations show limited and task-specific benefits when applied to severely compressed representations. While certain combinations can enhance performance on specific benchmarks, they cannot overcome the information bottleneck caused by using only a global image embedding.

\subsection{Performance of Auxiliary Objectives} \label{res:objective_functions}
\textbf{A pure next token prediction objective function achieves the best scores for our base model overall.} There are multiple potential causes for these results. First, the BLiMP benchmarks rely solely on linguistic information, therefore any auxiliary objective that competes with next token prediction can dilute the model’s focus on linguistic signals. This is reflected in the BLiMP score differences of 2.05\% and 4.06\% with the CLIP and LCG auxiliary functions, respectively. 

Second, the CLIP objective was
designed for a larger batch size than we could use with our computational budget and more data than the available samples in the image-caption dataset, which may have led to a limited impact. Third, the global image embeddings provide limited visual information, which seems to be insufficient to enable the contrastive auxiliary objectives to make fine-grained visual-linguistic alignments. 

Fourth, the alternation between text-only and image-caption epochs may cause training instability, since the auxiliary functions are only used during the image-caption epochs. Therefore, with a 10-epoch budget and the limited global image embeddings, there is no evident benefit of using contrastive learning auxiliary objectives for the benchmarks we selected.

From a cognitive perspective, these negative results may align better with theories of human language acquisition. Children do not learn language through explicit contrastive mechanisms where they simultaneously process what words do and do not mean across hundreds of examples, but rather in rich, multimodal contexts where meaning emerges from use rather than from explicit positive or negative examples. These results support our focus on architectural innovations, such as dynamic gating, which better capture the adaptive nature of human cognitive processing during language learning.

\section{Interpretability} \label{interpretability}
\begin{figure}
    \centering
    \includegraphics[width=\linewidth]{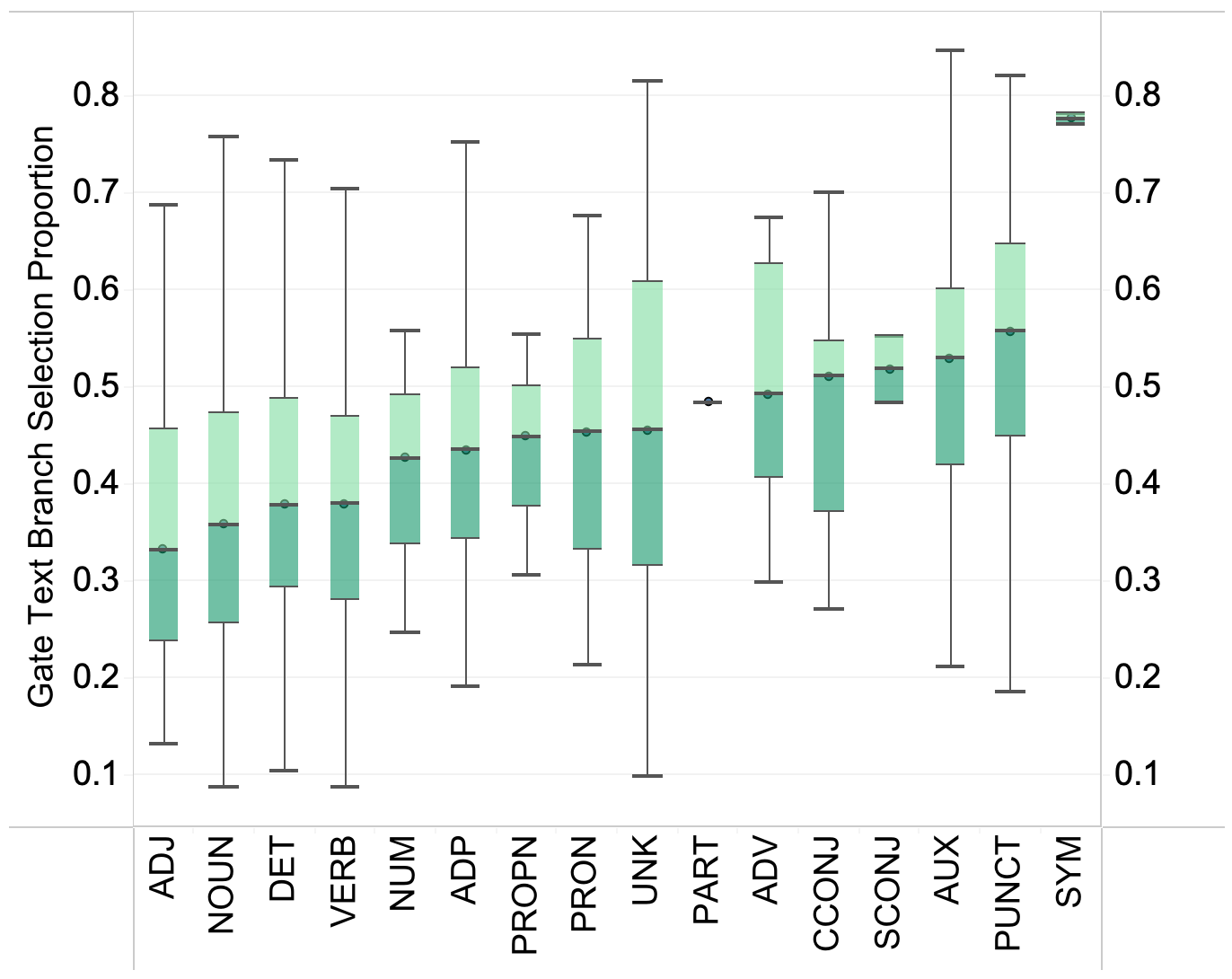}
    \caption{Our base model's aggregated gate values per part-of-speech for next token prediction, based on the Localized Narratives dataset. The parts-of-speech for each sentence were extracted using the spaCy English tagger \cite{honnibal2020spacy}. Lower values on the y-axis mean that the model attended less to the pure linguistic signals and more to the fused image-text representation when predicting the next token.\vspace{-15pt}}
    \label{fig:gate_selection}
\end{figure}

Figure \ref{fig:gate_selection} illustrates our model's gate value for next token prediction, aggregated per part-of-speech (PoS). The model is evaluated on held-out sentences from the Localized Narratives dataset, amounting to 1,034 tokens. The parts-of-speech for each sentence were extracted using the spaCy English tagger (\texttt{en\_core\_web\_sm}) \cite{honnibal2020spacy}.  The gate value plotted is the mean over the gate values per feature. A lower score means the model attended less to the pure linguistic signals and more to the fused image-text representation when predicting the next token.

There is an interpretable correlation between gate selection and part-of-speech. For the parts-of-speech which are open-class and generally more grounded (adjective, noun, proper noun, verb) \citep{haley2024grounded}, the model attends more to the image signals (left side of the plot), while for function words (conjunction, punctuation, symbols, auxiliary verbs, particles) the model attends more to the pure text. Furthermore, the model shows increased visual grounding for numerals, determiners and adpositions, suggesting they leverage visual information for counting and quantity (``two", ``three"), uniqueness (``a" vs ``the"), spatial reference (``this" vs ``that") and spatial relationships (``on", ``in", ``around"). We confirm the correlation between gate selection and parts-of-speech by running the Kruskal-Wallis statistical test \citep{mckight2010kruskal}, and obtaining  $(H = 154.91, p < 0.001)$.

We also find a statistically significant negative correlation between gate values and concreteness ($\rho = -0.139, p<0.001$) and imageability  ($\rho = -0.153,  p<0.001$) scores using the MRC Psycholinguistic Database \citep{coltheart1981mrc} and  Spearman's rank correlation test.

In Table \ref{tab:conc_imag_scores}, we illustrate the correlation between gate selection and concreteness by aggregating the gate values and defining meaningful categories based on score distributions from the MRC database using cutpoints at mean ± 1 SD. As shown, the correlation is weak ($|\rho|<0.2$), and the pattern is non-monotonic i.e., moderately abstract/concrete words show higher gate rates than the very abstract/concrete ones,
suggesting that other factors, such as part-of-speech, are more important in gating decisions. Similar results for imageability are available in Appendix \ref{appendix:imageability}.

\begin{table}[h]
\centering
\label{tab:gate_conc_by_bins}
\begin{tabular}{lll}
\toprule
 \textbf{Category} & \multicolumn{2}{l}{\textbf{Gate Selection}}  \\

 & Mean (SD) & \# words \\
\midrule

Very Abstract ($<$318) & 0.427 (0.141) & 420 \\
Abstract (318-438) & 0.471 (0.136) & 82  \\
Concrete (438-558) & 0.391 (0.155) & 80 \\
Very Concrete ($>$558) & 0.343 (0.139) & 119 9 \\

\bottomrule
\multicolumn{3}{l}{\footnotesize Categories defined as $\mu \pm \sigma$ based on the MRC database. \vspace{-5pt}} \\
\multicolumn{3}{l}{\vspace{-8pt}\footnotesize SD = standard deviation.} 
\end{tabular}
\caption{Mean gate value per concreteness bin for our base model (incorporating a soft gate per feature).}
\label{tab:conc_imag_scores}
\end{table}
\vspace{-10pt}
\section{Discussion}
\label{discussion}

\textbf{Missing modality problem.} In our experiments, we find that the split of the training data into text-only and image-caption datasets introduces complexity and instability during training. While we attempt to mitigate this by alternating epochs, the approach still yields performance oscillations (details in Appendices \ref {training_dynamics} and \ref{data_curriculum}). The BabyLM Challenge baselines address this problem by pairing text-only and image-caption in the same batch. However, this results in training the models on four times more image-caption samples than text-only samples. Moreover, to the best of our knowledge, there is no cognitive justification for this split.

\textbf{Data Curriculum.} We explore multiple data curriculum strategies that could optimise learning in our proposed framework. At a coarse-grained level, we explore different orderings between text-only and image-caption epochs. At a fine-grained level, we explore how mixing text-only and image-caption data within the same batch, either uniformly or non-uniformly, impacts the training dynamics and generalisation of the model. Empirical results suggest that alternating between image-caption and text-only epochs is the best strategy for our framework for the five benchmarks we selected in this work. A comprehensive discussion and results are available in Appendix \ref{data_curriculum}.

\textbf{Future work.} 
Based on the results and findings in this work, for the BabyLM Challenge Vision track, we make several observations for future work. First, the training dataset should be varied, with high-quality text that covers a range of English constructions. In particular, the training dataset should cover constructions (e.g., images paired with question-answers for VQA) and concepts (e.g., EWoK) present in the evaluation benchmarks. Second, for training stability and improved language acquisition, it may be more beneficial to train the model on a completely multimodal dataset, which is one promising avenue for future work. Third, given the limitations that the global image token introduced in this work, future work should use patch-token representations for the image input in order to enable richer multimodal learning -- that which is the aim of future iterations of this framework. Finally, it would be interesting to develop benchmarks that specifically reward cognitively-plausible mechanisms: i.e., evaluating the cognitive principles guiding the model's responses.

\section{Conclusion}
In this work, we show that token-wise dynamic gating enables small vision-language models to adaptively integrate linguistic and visual cues, yielding interpretable patterns and competitive performance under the BabyLM Challenge Vision track constraints. Our results highlight the promise of cognitively-inspired architectural design, while underscoring the need to address limitations in visual representation, training data and evaluation benchmarks to realise the full potential of multimodal learning in low-resource settings.

\section*{Limitations} \label{limitations}
Due to computational constraints, we could not use a larger batch size, which would have benefited the models trained with a contrastive loss objective \citep{chen2020simple}.

Two of the BabyLM Challenge benchmarks we used in this work showed limitations in evaluating our multimodal models, which potentially stem from a mismatch with the training data. 

Throughout our experiments, we find that EWoK demonstrates no sensitivity to changes in architecture or training strategy, with performance remaining around 50\% regardless of the experiment conditions. We therefore investigate the frequency of concepts tested by EWoK in the BabyLM Challenge training data, as previous research suggests that language models rely on frequency more than children do in word acquisition \citep{chang2022word}. A Regular Expression match for the concepts tested in EWoK over the training data revealed that in 37.69\% of the EWoK examples, at least one out of two concepts tested appears fewer than 100 times in the training data, with 13\% of test examples having both concepts appearing 0 times. Therefore, we conclude that the training dataset does not properly support EWoK evaluation.

By alternating between image-caption and text-only epoch, we also find score differences between epoch types for VQA (details in Appendix \ref{training_dynamics}). Our results suggest that VQA depends significantly on the presence of question-answer and turn-taking formats in the training dataset. This finding aligns with observations by \citet{laurenccon2024building}, who note that vision-language models typically only learn visual question answering during fine-tuning stages, not during pre-training, unless they are explicitly exposed to data following the VQA format. This is particularly problematic under the constraints of the BabyLM Challenge, where no fine-tuning stage exists, forcing models to acquire question-answering capabilities just from pre-training data that lacks examples similar to the VQA task.

As a general observation, training vision-language BabyLMs differs from training state-of-the-art large vision-language models, which rely on large pre-trained components. Moreover, VLMs can undergo multiple training stages where components are selectively frozen or unfrozen, higher-quality data is gradually introduced and the image resolution is progressively increased \citep{laurenccon2024building}. With limited data and a maximum of just 10 training epochs under the BabyLM Challenge constraints, implementing multi-stage training strategies becomes significantly more difficult.

\section*{Acknowledgments}
This paper reports on work supported by Cambridge University Press \& Assessment. It was performed using resources provided by the Cambridge Service for Data Driven Discovery (CSD3) operated by the University of Cambridge Research Computing Service, provided by Dell EMC and Intel using Tier-2 funding from the Engineering and Physical Sciences Research Council (capital grant EP/T022159/1), and DiRAC funding from the Science and Technology Facilities Council. 
We also particularly thank Dr Diana Galvan-Sosa.

\bibliography{custom}
\newpage
\appendix

\section{Dynamic Gating} \label{ch4:dynamic_gating}
We define four dynamic gating variants that operate at different granularity and decision levels: \textit{soft gate per feature}, \textit{soft gate per token}, \textit{hard gate per feature}, \textit{hard gate per token}.

\textbf{Input and Output.} All four versions of the dynamic gate have the same input and output. Let $h_\text{text}\in\mathbb{R}^{B\times T \times d_{\text{model}}}$ be the text hidden states after self-attention and $h_\text{crossAttn}\in\mathbb{R}^{B\times T \times d_{\text{model}}}$ be the output of the cross-attention between text and image, where $B$ is the batch size and $T$ is the sequence length. Then, the input to the dynamic gate is the concatenation of two hidden representations, $[h_\text{text};h_\text{crossAttn}]\in \mathbb{R}^{B\times T \times 2d_{\text{model}}}$. The output is represented by $h_\text{fused}\in\mathbb{R}^{B\times T \times d_{\text{model}}}$, which combined the pure linguistic representation with the visually-enriched representation based on the gating weights. In the case of a text-only input to the model, the dynamic gate module is skipped, and $h_\text{text}$ flows directly through the residual connection.

\begin{figure}
    \centering
    \includegraphics[width=1\linewidth]{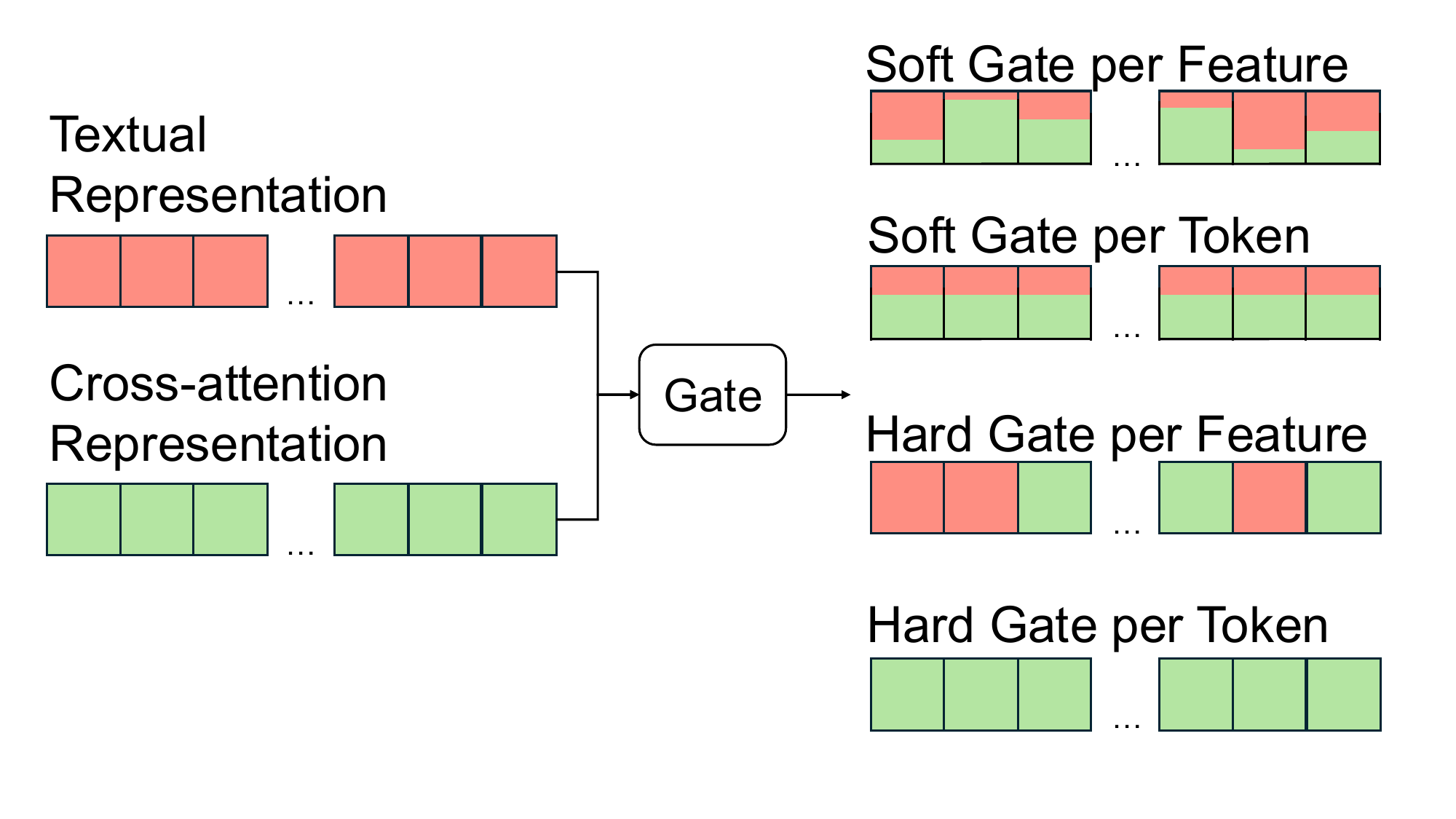}
    \caption{Conceptual output of different gating strategies for fusing textual and cross-attention representations. Each rectangular box represents a token, with the cells within representing dimensions. Red represents textual features, green represents cross-attention features and mixed colours represent fused features. Soft gates apply continuous weights, while hard gates make binary decisions, either per-feature (each dimension independently) or per-token (all dimensions together).}
    \label{fig:gate_architecture}
\end{figure}

Figure \ref{fig:gate_architecture} illustrates the conceptual output for each type of gate.
\subsection{Soft Gate per Feature}
This variant computes a continuous weight for each feature dimension $i\in\{0,...,d_{\text{model}} - 1\}$ using the sigmoid function. Concretely, the gate vector is computed as:
\begin{equation}
  g = \sigma\!\bigl(\text{Linear}\,[\,h_{\text{text}};\,h_{\text{crossAttn}}\,]\bigr)\!\in\![0,1]^{B \times T \times d_{\text{model}}}
\end{equation}

The dynamically fused representation is then calculated as:
\begin{equation}
  h_{\text{fused}}
  = g \,\odot\, h_{\text{text}}
    + (1 - g) \,\odot\, h_{\text{crossAttn}}
\end{equation}

We use this variant of dynamic gating in the base model.

\subsection{Soft Gate per Token}
The soft gate per token calculates a single continuous weight (a scalar), which we apply to all features in the hidden representations:
\begin{equation}
\begin{split}
  g = \sigma\!\Bigl(\text{Linear}\bigl([\;h_{\text{text}};\,h_{\text{crossAttn}}\bigr])\Bigr)
     \in[0,1]^{B \times T \times 1}, \\
  h_{\text{fused}} 
  = g \,\odot\, h_{\text{text}}
     + \bigl(1 - g\bigr) \,\odot\, h_{\text{crossAttn}}
\end{split}
\end{equation}

\subsection{Hard Gate per Feature} \label{subsection:hard_gate_per_feature}
Drawing inspiration from \citet{xue2023dynamic}, we extend the soft gating variants to a hard selection mechanism using the Gumble-Softmax reparametrisation trick \citep{jang2016categorical}.

The hard gate per feature variant enforces each dimension to choose completely between linguistic or visually-enriched representations. We first compute a 2-way discrete choice for the two hidden representations using a linear layer:
\begin{equation}
  g\!=\!\text{Linear}\!([\,h_{\text{text}};\,h_{\text{crossAttn}}\,])
  \!\in\![0,1]^{B \times T \times d_{\text{model}}\times2}
\end{equation}

Each pair of logits $(l_{b,t,i,0}, l_{b,t,i,1})$ corresponds to the scores for ``use $h_\mathrm{text}$" versus ``use $h_\mathrm{crossAttn}$" for feature $i$ at position $(b,t)$. A straightforward hard gate would then be:
\begin{equation}
  g_{b,t,i}
  = \arg\max\bigl(l_{b,t,i,0},\,l_{b,t,i,1}\bigr)
\end{equation}

However, since $g$ is a one-hot vector, it is not differentiable. Therefore we employ a soft gate $\tilde{g}$ during training using Gumble-Softmax, similar to \citet{xue2023dynamic}, to enable back-propagation:
\begin{equation}
  \tilde l_{b,t,i,j}
  = \frac{l_{b,t,i,j} + z_{b,t,i,j}}{\tau},
  \;
  z_{b,t,i,j} \sim \text{Gumbel}(0,1)
\end{equation}

where  $\tau$ is the Softmax temperature. We then apply Softmax over the two classes and select the probability corresponding to $h_\mathrm{text}$ as the soft gate:
\begin{equation}
  y_{b,t,i,j}
= \frac{\exp\bigl(\tilde l_{b,t,i,j}\bigr)}
         {\sum_{k=0}^1 \exp\bigl(\tilde l_{b,t,i,k}\bigr)},
  \;
  y \!\in\![0,1]^{B \times T \times d_{\text{model}} \times 2}
\end{equation}
\begin{equation}
  \tilde g_{b,t,i} = y_{b,t,i,0},
  \;
  \tilde g \;\in\; [0,1]^{B \times T \times d_{\text{model}}}
\end{equation}

$h_\text{fused}$ is then be computed as:
\begin{equation}
  h_{\text{fused}}
  = \tilde g \,\odot\, h_{\text{text}}
    + \bigl(1 - \tilde g\bigr)\,\odot\, h_{\text{crossAttn}}
\end{equation}

During training, we anneal the Softmax temperature $\tau$ from 1.0 to 0.1 over 80\% of the training steps of an image-caption epoch, gradually transitioning from soft to nearly discrete selection. During inference, we convert $\tilde{g}$ to a true one-hot gate $g$ using $\arg\max$.

\subsection{Hard Gate per Token}
In the per token variant of the hard gate, we collapse the feature-wise gate into a single binary decision. The calculations, training and inference remain the same as in subsection \ref{subsection:hard_gate_per_feature}, yet the shape of the parameters changes. The summary of the calculations in this variant is as follows:
\begin{equation}
  l
  = \text{Linear}\bigl([\;h_{\text{text}};\,h_{\text{crossAttn}}\bigr])
  \;\in\;\mathbb{R}^{B \times T \times 2}
\end{equation}
\begin{equation}
  y
  = \text{GumbelSoftmax}\bigl(l,\,\tau\bigr)
  \;\in\;\mathbb{R}^{B \times T \times 2}
\end{equation}
\begin{equation}
  \tilde g_{b,t}
  = y_{b,t,0}
  \;\in\;[0,1]
\end{equation}
\begin{equation}
  h_{\text{fused}}
  = \tilde g \,\odot\, h_{\text{text}}
    + \bigl(1 - \tilde g\bigr)\,\odot\, h_{\text{crossAttn}}
\end{equation}

where the scalar $\tilde{g}$ is broadcasted over all $d_\text{model}$ features.

\section{Feature Representation} \label{ch4:feature_representation}
\subsection{Feature-wise Linear Modulation (FiLM)}
To address the limited representational capacity of a single image token, we incorporate Feature-wise Linear Modulation (FiLM) \citep{perez2018film} as an intra-modal conditioning mechanism. FiLM modulates neural network features through a feature-wise affine transformation, enabling one modality or context to dynamically influence another. Specifically, it applies scaling and shifting to a feature map based on a conditioning input, and can be easily implemented in transformers as follows:

Let $h_{m_{1}}, h_{m_{2}} \in \mathbb{R}^{B \times T \times d_\text{model}}$ be hidden state feature representations with $m_1$ indicating the primary features and $m_2$ the conditioning features, $m_1 \neq m_2$. Then,
\begin{equation}
  \text{FiLM}\bigl(h_{m_1},\,h_{m_2}\bigr)
  = \gamma \,\odot\, h_{m_1} + \beta
\end{equation}
where $\gamma, \beta \in\mathbb{R}^{B \times T \times d_\text{model}}$ are scaling and shifting parameters predicted by linear layers from $h_{m_{2}}$.

\subsection{Dynamic Intra Modulation (DyIntra)}
Alternatively to FiLM, we explore the DyIntra module proposed in \citep{gao2019dynamic}, a scaling mechanism that modulates primary features using conditioning features via a simple gating mask.  DyIntra predicts a positive‐only gain for each representation, allowing it to boost its own hidden features based on cross‐modal context without shifting. 

Formally, let $h_{m_{1}}, h_{m_{2}} \in \mathbb{R}^{B \times T \times d_\text{model}}$ be hidden state feature representations, $m_1 \neq m_2$. Then, DyIntra computes:
\begin{equation}
  m
  = \sigma\!\bigl(\text{Linear}(h_{m_{2}})\bigr)
  \;\in\;[0,1]^{B \times T \times d_{\text{model}}}
\end{equation}
\begin{equation}
  \text{DyIntra}\bigl(h_{m_1},\,h_{m_2}\bigr)
  = (1 + m)\,\odot\,h_{m_1}
\end{equation}
\paragraph{Choosing $m_{1}$ and $m_{2}$.} There are several points in our base model where we could integrate a FiLM or DyIntra modulation module. We evaluate and motivate three such choices as follows:
\begin{enumerate}
    \item $m_{1} =$ self-attention (output), $m_{2} =$ image: modulating the text self-attention output with visual features may allow the model to adjust how text tokens relate to each other based on visual context;
    \item $m_{1}$=cross-attention (output), $m_{2} =$ image: modulating cross-attention features with the original image may refine the vision-language fusion by emphasising features that align with the global visual representation;
    \item $m_{1} =$ image, $m_{2} =$ text: modulating image features based on textual context may allow the model to dynamically highlight relevant visual information for the current linguistic processing needs.
\end{enumerate}
\subsection{Channel Attention}
To implement channel attention for only one image token, we use the Excitation formula from the Squeeze-and-Excitation method \citep{hu2018squeeze} as follows:
\begin{equation}
\begin{split}
  h_{\text{image}}'
  = \sigma\!\Bigl(W_{2}\,\text{ReLU}\bigl(W_{1}\,h_{\text{image}}\bigr)\Bigr)
    \,\odot\,h_{\text{image}},\\ 
    W_{1}\in\mathbb{R}^{(d_\text{model}/r \times d_\text{model})}, \; W_{2}\in\mathbb{R}^{(d_\text{model} \times d_\text{model}/r)}
\end{split}
\end{equation}
where $h_{image}$ is the output of the image encoder and $r = 16$ is the reduction ratio. We expect this method to help the model focus on the most informative features of the visual embedding, improving the quality of image representations.

\section{Auxiliary Objective Functions} \label{ch4:objective_functions}
\subsection{Contrastive Language-Image Pre-training (CLIP)}
We incorporate the Contrastive Language-Image Pre-training (CLIP) objective function into the training of our base model for image-caption epochs steps, as follows:

For each sample in a batch, we first extract pooled representations from both image and text modalities. For text,we compute mean pooling over the the output of the text embedding module, which we denote $t_\text{pooled}$. For the image, we use the output of the image encoder directly as its length is 1, $i_\text{pooled}$. Both $t_\text{pooled}$ and $i_\text{pooled}$ are then projected to a shared contrastive embedding space through specific linear projection layers. We L2-normalise both representations before computing similarity scores. The contrastive loss is formulated as a bidirectional InfoNCE objective \citep{oord2018representation} with learnable temperature $\tau$. It combines text-to-image and image-to-text matching losses, where each direction maximises the similarity between matched pairs while minimising similarity with all other pairs in the batch. The final loss is computed as $\mathcal{L}_{\text{contrastive}} = \frac{1}{2}(\mathcal{L}_{\text{t2i}} + \mathcal{L}_{\text{i2t}})$.

The complete training objective then becomes:
\begin{equation}
 \mathcal{L}_\text{total} = \mathcal{L}_\text{NTP} + \lambda \mathcal{L}_\text{contrastive}   
\end{equation}

where $\lambda$ represents the weight of the contrastive loss and $NTP$ stands for \textit{next-token prediction}.

In our experiments, we initialise $\tau$ to 0.07 and constraint it between 0.05 and 1 during training for stability, and set $\lambda$ to 1.

\subsection{LexiContrastive Grounding (LCG)}

LexiContrastive Grounding (LCG) \citep{zhuang2024lexicon} is a training procedure that implements a grounded language learning objective similar to CLIP. While CLIP operates at sentence level, LCG computes similarity scores at the word level. To calculate the cross-modality contrastive learning loss, LCG leverages the first hidden layer of a model, which stores lexical information. The authors also limit the attention mask applied to the first layer to a previous two-word window in order to encode less linguistic context. The contrastive loss is then calculated per batch from all the token-level representations outputted by the first layer. 

For our model, we adapt and implement the LCG during the image-caption epochs as follows:

Let $(\text{text}_i, \text{image}_i)$ represent the image-caption pairs in a batch, where $i \in \{1, 2, \ldots, n\}$ and $n$ is the batch size. Each caption $\text{text}_i$ contains $m_i$ tokens: $(t_{i,1}, t_{i,2}, \ldots, t_{i,m_i})$.

To obtain lexically-focused representations, we extract the textual representation from the first layer after the residual connection applied to self-attention:
\begin{equation}
  h_{1}\bigl(\text{text}_{i}\bigr)
  = \text{text}_{i}
    + \text{SelfAttn}\!\bigl(\text{LayerNorm}(\text{text}_{i})\bigr)
\end{equation}
In our implementation, we  experimented with applying a narrow two-word attention mask, however, we noticed conflicts with the next token prediction loss. Specifically, applying the two-word attention mask in the first layer was preventing the next token prediction loss from decreasing. We tried applying the two-word attention mask solely to extract the hidden representation, then switching to the original causal mask for the rest of first layer's forward pass, as well as skipping the cross-modal fusion in the first layer, but neither approach fixed the problem. Therefore, we decided to use the standard causal attention mask when extracting the first layer textual hidden representation, as ablation studies in the original research \citep{zhuang2024lexicon} did not indicate a significant loss in performance for this case.

Let $h_1(\text{text}_i, j) \in \mathbb{R}^{d_\text{model}}$ be the first layer representation of the $j$-th token (the $j$-th row of the matrix) in the $i$-th caption and $\text{enc}(\text{image}_i) \in \mathbb{R}^{d_\text{model}}$ represent the output of the image encoder for the $i$-th image. Then, the matching score between the $j$-th token in the caption $k$ and image $i$ is calculated as:
\begin{equation}
\small s(i,j,k)\!=\!\frac{\bigl(\,M_{\text{image}}\!\cdot\!\text{enc}(\text{image}_{i})\bigr)^{T}\!\cdot\!\bigl(M_{\text{text}}\!\cdot\!h_{1}(\text{text}_{k},\,j)\bigr)}
         {\tau}
\end{equation}
where $M_\text{image}, M_\text{text} \in \mathbb{R}^{d_\text{model} \times d_\text{model}}$ are learned projection matrices and $\tau$ is a learnable temperature parameter, which we clamp between $[0.05, 2.0]$ for training stability.

For each valid token position, we then compute the LCG contrastive learning loss as:
\begin{equation}
  \mathcal{L}_{\text{LCG}}
  = \sum_{i=1}^{n} \sum_{j=1}^{m_i}
    \mathbf{1}_{\text{valid}}(i,j)
    \;\cdot\;\frac{1}{2}\bigl[\ell_1(i,j) + \ell_2(i,j)\bigr]
\end{equation}
where $\mathbf{1}_\text{valid}(i,j)$ is an indicator function for non-padded tokens, and:
\begin{equation}
  \ell_{1}(i,j)
  = \frac{e^{s(i,j,i)}}{\sum_{k=1}^{n}e^{s(k,j,i)}},
  \;
  \ell_{2}(i,j)
  = \frac{e^{s(i,j,i)}}{\text{neg}(i,j)}
\end{equation}
The negative term $\text{neg}(i,j)$ is defined as:
\begin{equation}
  \text{neg}(i,j)
  = e^{s(i,j,i)}
    + \sum_{\substack{k=1 \\ k \neq i}}^{n}
      \sum_{o=1}^{m_k}
        \mathbf{1}_{\text{valid}}(k,o)\,\cdot\,e^{s(i,o,k)}
\end{equation}
The total loss combines next-token prediction with word-level contrastive learning loss:
\begin{equation}
\mathcal{L}_\text{total} = \mathcal{L}_\text{NTP} + \lambda \cdot \mathcal{L}_\text{LCG}
\end{equation}

where $\lambda$ is a hyperparameter controlling the strength of visual grounding. We set $\lambda$ to 1 through trial and error such that $\mathcal{L}_\text{NTP}$ and $\mathcal{L}_\text{LCG}$ have the same magnitude.

We use auxiliary functions only during the image-caption epochs, as the image processing stream is skipped for text-only samples.

\section{Evaluation Benchmarks and Training Data}\label{appendix:background_evaluation_pipeline}

The evaluation pipeline of the BabyLM Challenge consists of both text-only and multimodal benchmarks. To evaluate our models, we use the following benchmarks from the BabyLM Challenge:
\begin{itemize}
    \item BLiMP (The Benchmark of Linguistic Minimal Pairs) \citep{warstadt2020blimp} evaluates the linguistic abilities of language models through grammatical acceptability judgements. It consists of minimal pairs of sentences testing a specific phenomenon in syntax, semantics or morphology. Each pair contains one well-formed sentence and one ungrammatical sentence. Models are evaluated by checking whether they assign a higher probability to the grammatical sentence in each pair. 
    \item BLiMP Supplement is a held-out evaluation set introduced in the BabyLM Challenge, consisting of five additional linguistic tasks. 
    \item Elements of World Knowledge (EWoK) \citep{ivanova2024elements} is a zero-shot benchmark that targets specific world concepts such as social interactions, spatial relations and physical dynamics. It uses minimal pairs of context-target combinations, where the same target sentence is plausible given one context but implausible given another. Models are evaluated by checking whether they assign a higher probability to the correct context-target pair.

\item Winoground \citep{thrush2022winoground} evaluates visio-linguistic compositional reasoning in vision-language models. The dataset consists of hand-curated examples where models must correctly match two images with two captions that contain identical words but in different orders (e.g., ``some plants surrounding a lightbulb" vs\ ``a lightbulb surrounding some plants"). Models are evaluated by checking whether they assign a higher probability to the correct caption given the input image.

\item VQA v2.0 \citep{goyal2017making} is an evaluation dataset containing pairs of similar images with identical questions but different correct answers, which forces models to ground their responses in visual content rather than rely on linguistic priors alone. Questions cover multiple categories, such as object recognition, counting and spatial reasoning. Models are evaluated based on which answer they assign the highest probability given the input image and question.
\end{itemize}

The BabyLM Challenge organisers provide an image-text pre-training dataset for the Vision track, which we use in our work. This dataset consists of two parts: text-only data and text-image data, each containing approximately 50 million words. 

The text-only dataset is a subset of the training data proposed for the BabyLM Challenge text-only track. The organisers argue that this dataset is cognitively plausible, consisting of child-directed speech (CHILDES \citep{macwhinney2000childes}), dialogue (British National Corpus (BNC) conversation section\footnote{\url{http://www.natcorp.ox.ac.uk}}, Switchboard Dialog Act Corpus \citep{stolcke2000dialogue}), children's stories (Project Gutenberg \citep{gerlach2020standardized}), movie subtitles (OpenSubtitles \citep{lison2016opensubtitles2016}) and Wikipedia\footnote{\url{https://www.wikipedia.org}} content. 

The multimodal dataset consists of image-caption pairs selected from the Conceptual Captions 3M dataset \citep{sharma2018conceptual}, and the MS-COCO \citep{lin2014microsoft} and Open Images \citep{kuznetsova2020open} subsets of the Localized Narratives dataset \citep{pont2020connecting}. The Conceptual Captions dataset consists of millions of images paired with natural language descriptions automatically scraped, cleaned and filtered from web image alt-text, while the Localized Narratives dataset contains image-caption pairs manually annotated with synchronised mouse traces that spatially ground each word or phrase to specific regions in the image. The images are provided in both raw format and as visual embeddings computed by a visual model using DINOv2 \citep{babylm_call_for_papers_2, oquab2023dinov2}, a state-of-the-art unsupervised learning algorithm. We use these visual embeddings in both our training and evaluation due to computational constraints.

\section{Data Curriculum} 
\label{data_curriculum}

Since the training dataset consists of both text-only data and image-caption data, each accounting for 50M words, we implement and analyse multiple coarse-grained and fine-grained data curriculum strategies for training.

\textbf{Coarse-grained epochs}: We load the text-only and image-caption data in separate PyTorch \citep{imambi2021pytorch} data loaders, where each data loader alone is used for one epoch. For the 10 epochs constraint of the BabyLM Challenge, this results in 10 text-only epochs and 10 image-caption epochs. We then experiment with the following:
    \begin{enumerate}
        \item Alternating between image-caption epochs and text-only epochs;
        \item Training on all text-only epochs first, then on the image-caption epochs;
        \item Training on all image-caption epochs first, then on the text-only epochs.
    \end{enumerate}
    
\textbf{Fine-grained epochs}: For the fine-grained epochs, we define the following two training strategies:
    \begin{enumerate}
        \item We load both the text-only data and the image-caption data in the same data loader, where we pair the text-only data with image tensors filled with 0s for uniformity. The cross-modality path is still skipped in the text-only samples. The original text data is provided in \textit{.txt} files, and we process each text line as one sample. For the image-caption data, we process each (image, caption) pair as one sample. In this setting, the text-only data has twice as many samples as the image-caption data. Therefore, loading and shuffling them in the same data loader results in a non-uniform distribution between the two and more unstable training.
        \item For a uniform distribution between the image-caption data and the text-only data, we take inspiration from the GitHub repository\footnote{\url{https://github.com/aaronmueller/babylm_multimodal_training}} used to train the BabyLM 2024 Challenge baselines, where the authors pair each text-only input with one image-caption input in the same batch sample, resulting in uniform batches. Therefore, in one training step, we perform two forward passes: one using the text-only input and one using the image-caption input. We then sum the losses from each pass and do one backward propagation using the total loss. However, since there are twice as many text-only samples than image-caption samples, this results in training the model twice on the image-caption dataset. For 10 training epochs, this equals 10 text-only epochs and 20 image-caption epochs. 
    \end{enumerate}

For a fair comparison among all of the methods we implement in this work, we alternate between text-only epochs and image-caption epochs in our experiments exploring architectural changes and auxiliary objective functions. That is because the contrastive learning objective functions compute similarity scores between a caption and all the images in a batch. If the batch contains (many) text-only samples, it cancels the effect of the auxiliary losses. 

\begin{figure*}[h!]
    \centering    
    \begin{subfigure}[b]{0.49\linewidth}
        \centering
        \includegraphics[width=\textwidth]{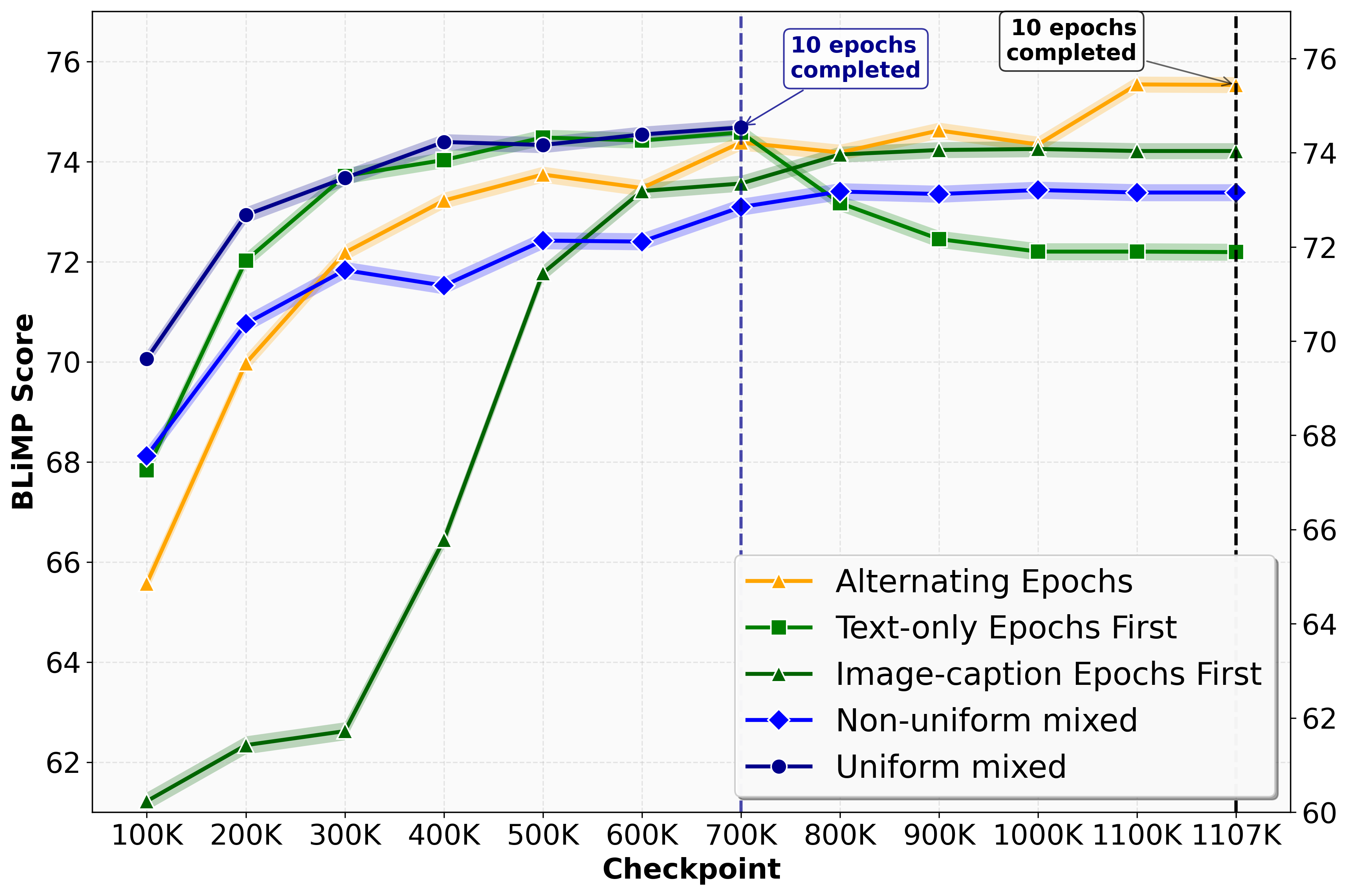}
        \caption{BLiMP scores.}
        \label{fig_data:sub1}
    \end{subfigure}
    \hfill
    \begin{subfigure}[b]{0.49\textwidth}
        \centering
        \includegraphics[width=\textwidth]{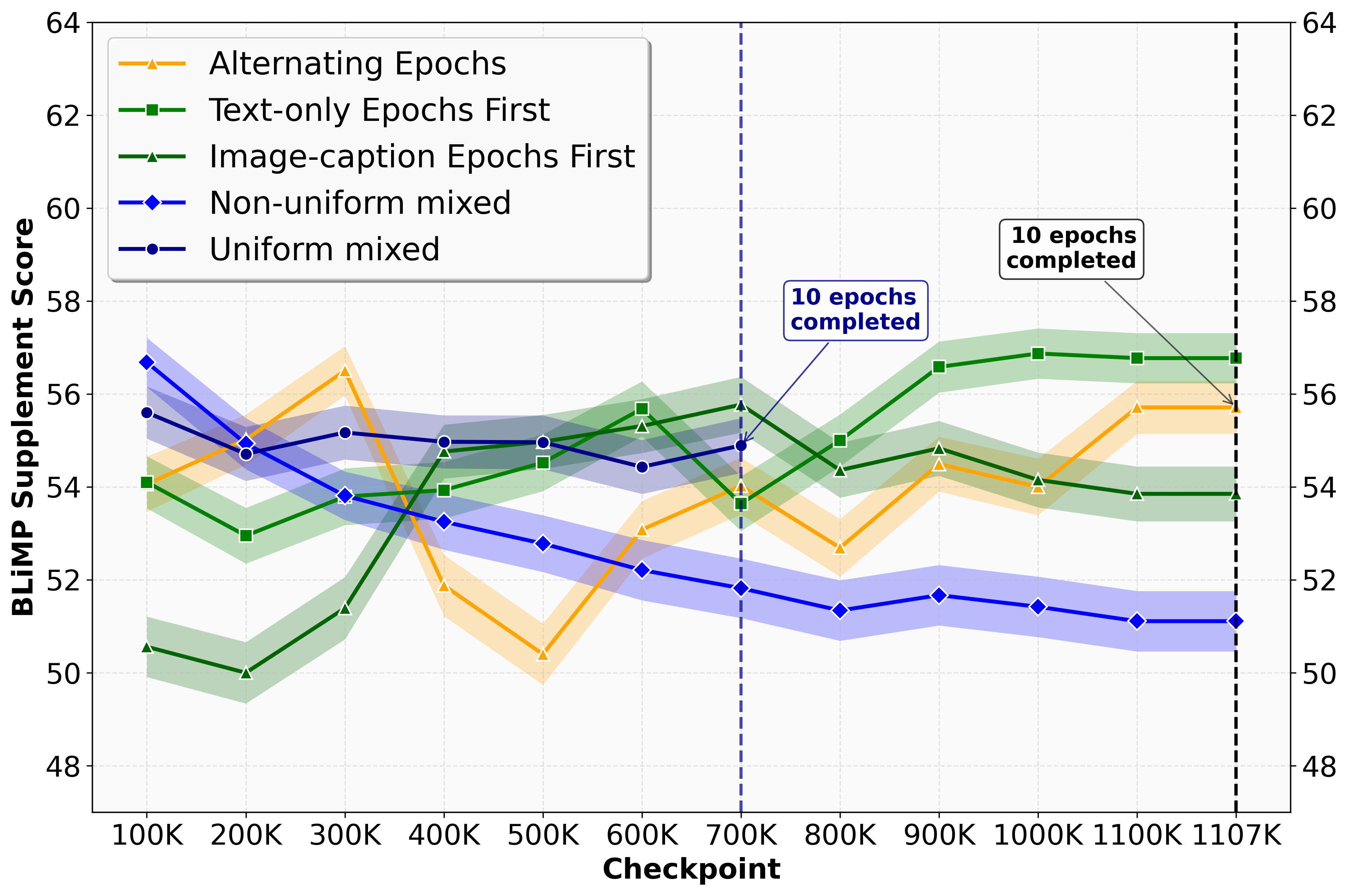}
        \caption{BLiMP Supplement scores.}
        \label{fig_data:sub2}
    \end{subfigure}
    
    \vspace{0.5em} 
    
    \begin{subfigure}[b]{0.49\textwidth}
        \centering
        \includegraphics[width=\textwidth]{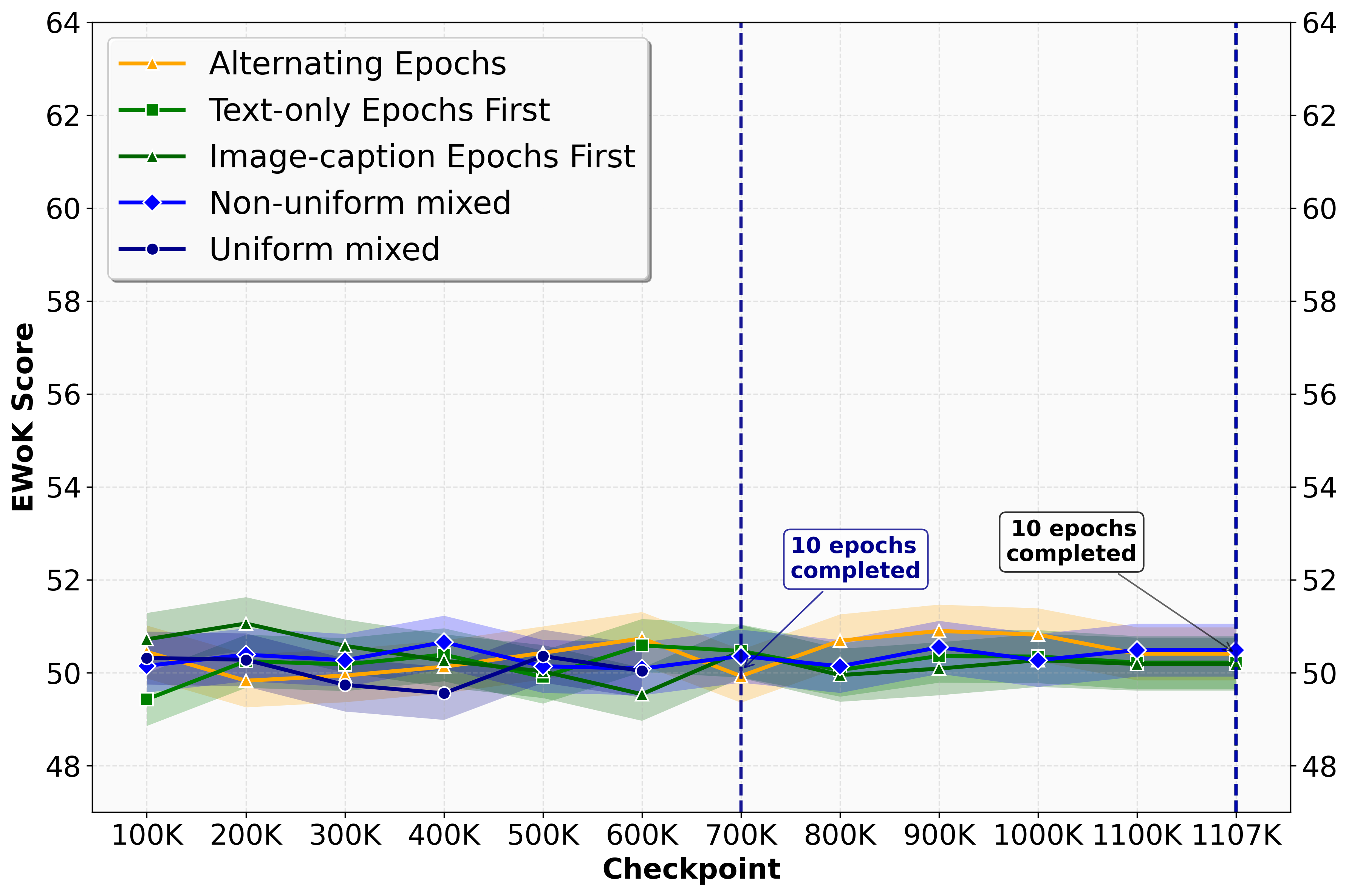}
        \caption{EWoK scores.}
        \label{fig_data:sub3}
    \end{subfigure}
    \hfill
    \begin{subfigure}[b]{0.49\textwidth}
        \centering
        \includegraphics[width=\textwidth]{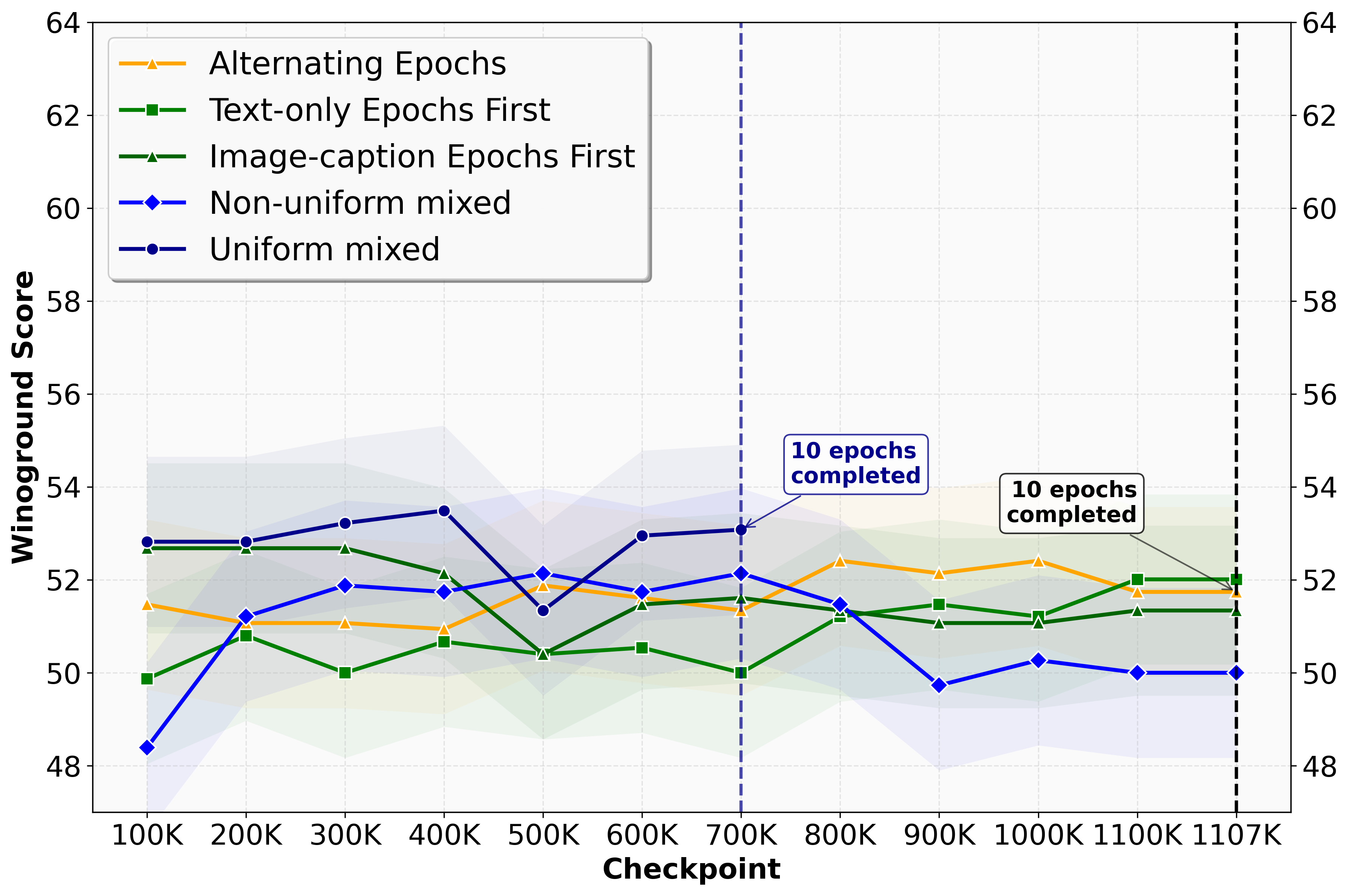}
        \caption{Winoground scores.}
        \label{fig_data:sub4}
    \end{subfigure}
    \hfill
    \begin{subfigure}[b]{0.49\textwidth}
        \centering
        \includegraphics[width=\textwidth]{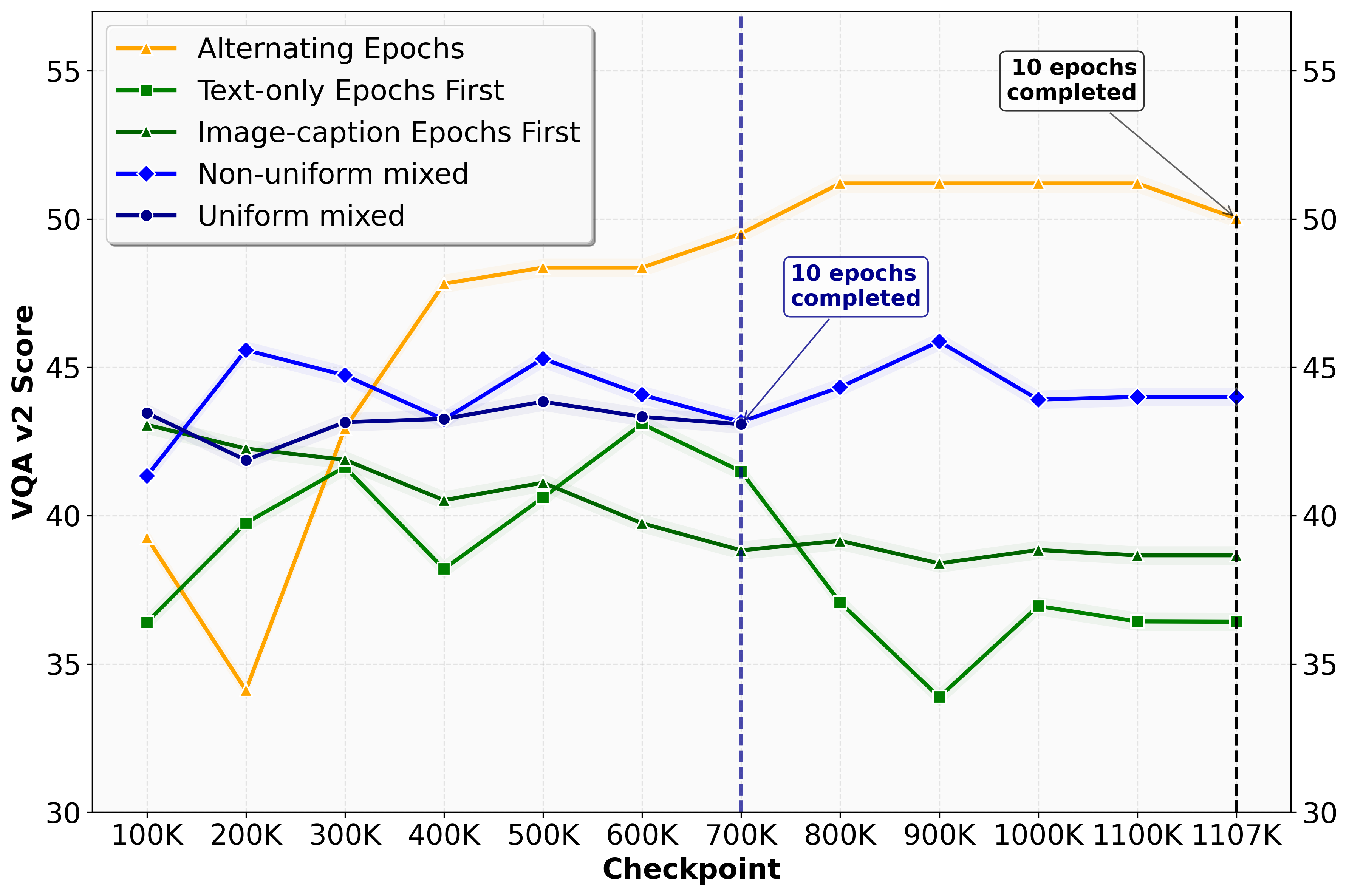}
        \caption{VQA scores.}
        \label{fig_data:sub5}
    \end{subfigure}
  \caption{The performance of our base model on BLiMP, BLiMP Supplement, EWoK, Winoground and VQA for different data curriculum strategies, evaluated using the BabyLM Challenge 2024 evaluation pipeline. The \textit{uniform mixed} strategy follows a different definition than the others, where the number of steps in an epoch equals the number of text-only samples. This results in $\sim$700K training steps for 10 epochs, which are marked in the graphs by the blue dashed line. The end of 10 epochs for the other data curriculum strategies is marked by the black dashed line at step $\sim$1107K.}
    \label{fig:data_curriculum}
\end{figure*}

Empirical results we obtain support this choice among the data curriculum strategies we define. Figure \ref{fig:data_curriculum} visualises the scores of our base model for the different data curriculum strategies evaluated using the BabyLM Challenge 2024 evaluation pipeline\footnote{\url{https://github.com/babylm/evaluation-pipeline-2024}}. 

For BLiMP, the pattern in subfigure \ref{fig_data:sub1} suggests that (1) the text-only dataset supports the BLiMP benchmark far more than the image-caption one, and (2) these strategies result in catastrophic forgetting for the model by the end of training.

For BLiMP Supplement, the optimal data curriculum strategy is less clear, as the model's performance oscillates when alternating between epoch types. Comparing the \textit{text-only epochs first} and \textit{image-caption epochs first} strategies shows that the image-caption dataset better supports the model on BLiMP Supplement than the text-only dataset. Interestingly, the model's performance score consistently decreases over checkpoints when the model is trained using the \textit{non-uniform mixed} strategy. A possible explanation for this result is that since there are more text-only samples in a batch than image-caption samples, the gradient updates are dominated by the text-only data, reducing the effect of the image-caption samples. 

Similar to the other analyses in this work, changing the data curriculum strategy has no visible effect on the EWoK benchmark, underscoring that the training data might not be well-suited for this benchmark. 

Training our base model using the \textit{uniform mixed} strategy results in a higher Winoground score, with several checkpoints achieving over 53\% on this benchmark. However, a significant factor contributing to this result is the amount of image-caption training data, which is double for this strategy than for the others. Comparing the \textit{text-only epochs first} and \textit{image-caption epochs first} strategies, it can be seen that the model performs better on Winoground when consistently trained on the image-caption dataset. Using the \textit{non-uniform mixed} strategy results in a more unstable performance and a lower final score, possibly due to the dominance of text-only samples in the training batches.

As shown in subfigure \ref{fig_data:sub5}, the alternating between text-only and image-caption epochs strategy achieves the best performance on VQA. There is a significant performance gap between the model trained using \textit{alternating epochs} compared to the \textit{mixed} strategies (over 5\%), as well as the \textit{text-only epochs first} and \textit{image-caption epochs first} strategies (over 10\%). The \textit{alternating epochs} strategy shows an almost consistent increase over checkpoints, whereas the model's performance in the \textit{mixed} variants remains flat, and decreases for the \textit{text-only epochs first} and \textit{image-caption epochs first} strategies. The results of the coarse-grained strategies are likely due to the training data. The image-caption dataset supports the visual reasoning component of VQA, while the text-only dataset supports the question format by containing turn-taking constructions and a significantly larger number of questions than the other dataset. Training the model consistently on only one epoch type deprives it of one of these complementary components. Training by alternating between epoch types appears to strike a balance and avoid catastrophic forgetting. The results of the \textit{uniform mixed} strategy are slightly surprising given that the Flamingo and GIT baselines achieve higher VQA scores using this approach, however, the difference could stem from using a lower learning rate and a different text-only to image-caption data ratio.

\section{Training Dynamics} \label{training_dynamics}
\begin{figure*}[h!]
    \centering
    \begin{subfigure}[b]{0.75\textwidth}
        \centering
        \includegraphics[width=\textwidth]{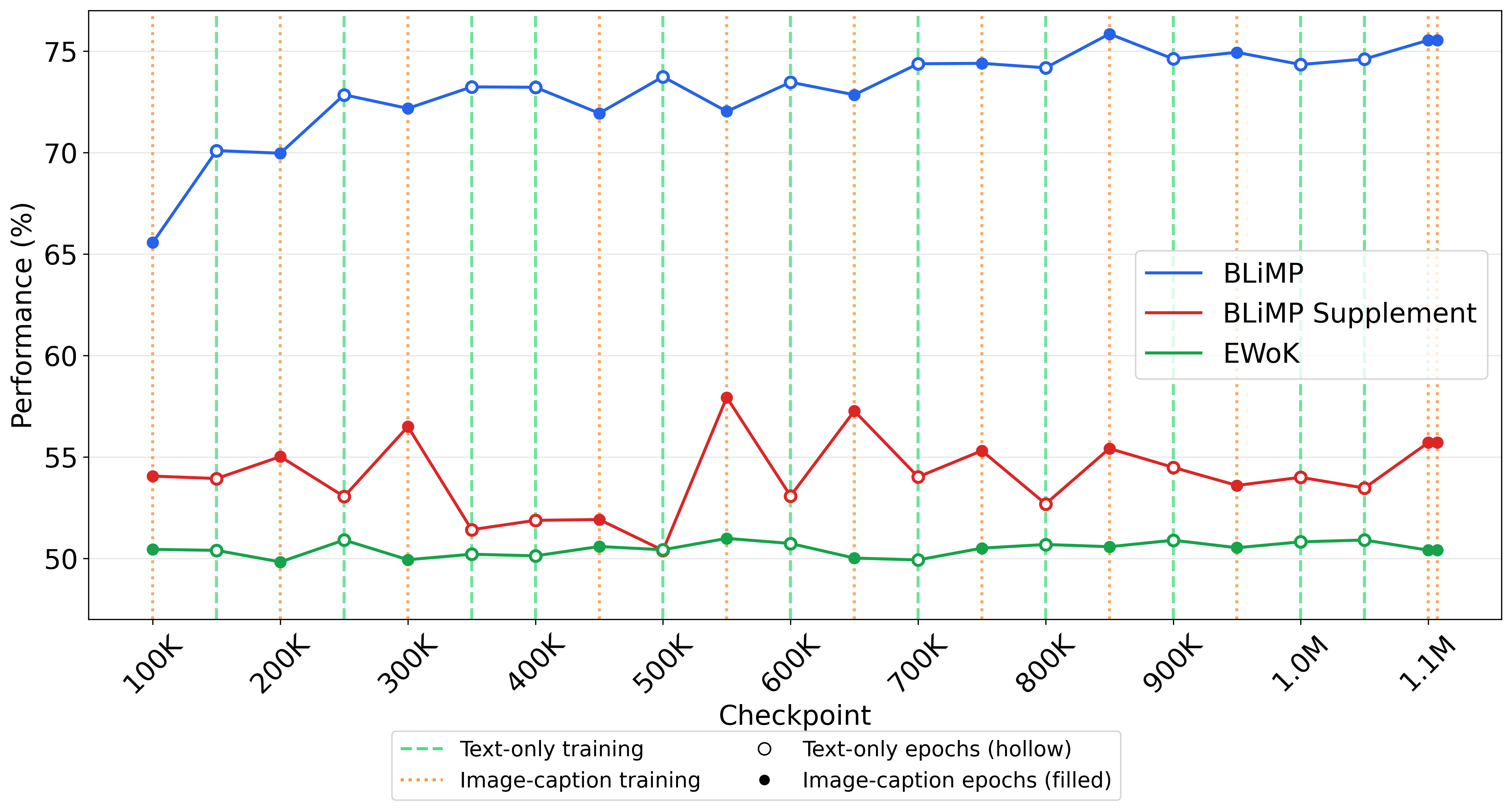}
        \caption{BLiMP, BLiMP Supplement and EWoK scores.}
        \label{training_dyn:sub1}
    \end{subfigure}
    \hfill
    \begin{subfigure}[b]{0.75\textwidth}
        \centering
        \includegraphics[width=\textwidth]{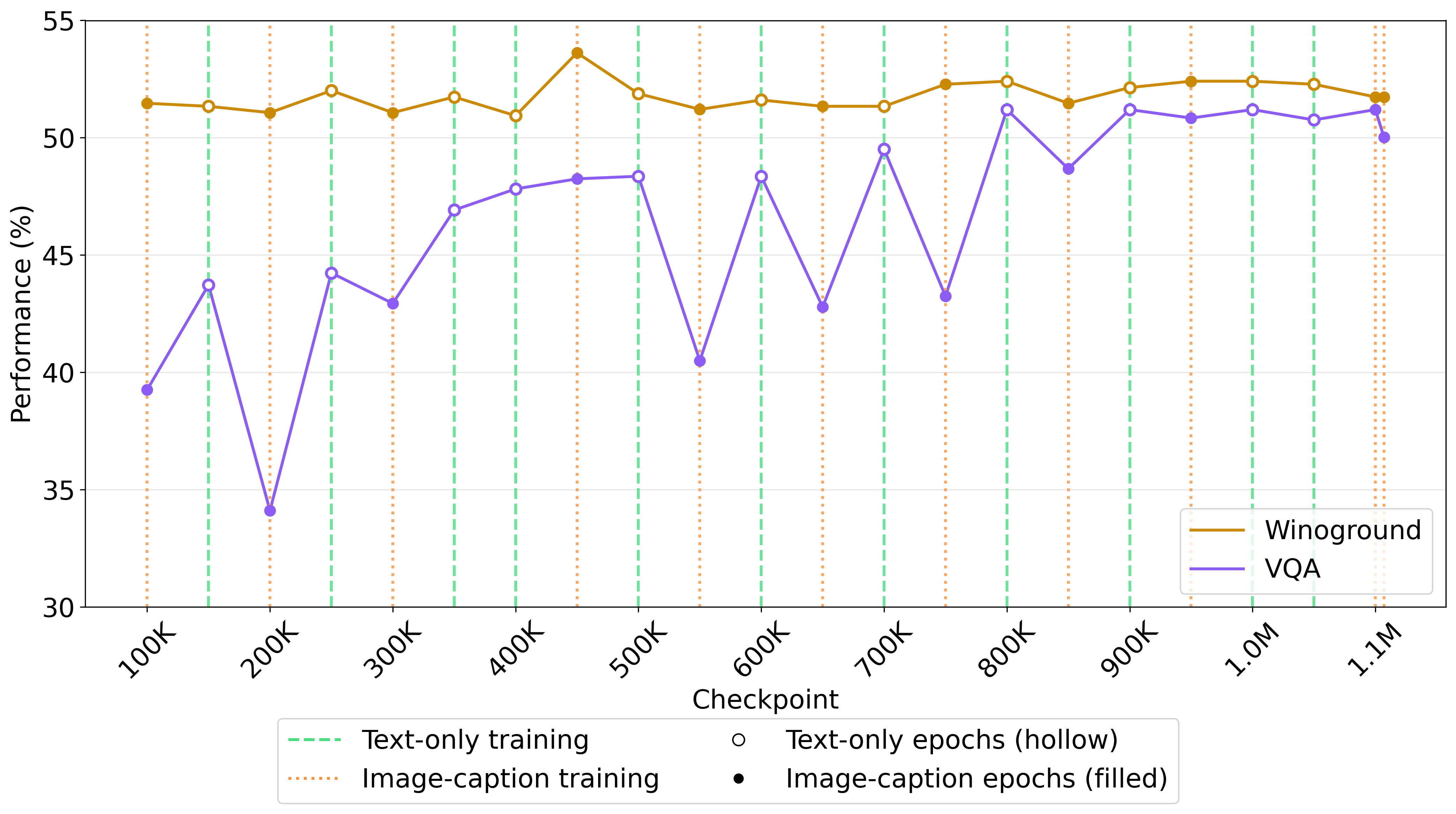}
        \caption{Winoground and VQA scores.}
        \label{training_dyn:sub2}
    \end{subfigure}
       
    \caption{The performance of our base model every 50,000 steps on the BLiMP, BLiMP Supplement, EWoK, Winoground and VQA benchmarks. The brown doted lines indicate that the checkpoint was saved during a text-only epoch, while the green dashed lines indicate that the checkpoint was saved during an image-caption epoch. \vspace{-0.3cm}}
    
    \label{fig:training_dyn}

\end{figure*}
In figure \ref{fig:training_dyn}, we visualise the performance of our base model, evaluated every 50,000 steps on the five benchmarks using the BabyLM Challenge 2024 evaluation pipeline, when alternating between text-only and image-caption epochs. The brown dotted lines indicate that the checkpoint was saved during a text-only epoch, while the green dashed lines indicate that the checkpoint was saved during an image-caption epoch. For BLIMP Supplement and VQA, an interesting pattern emerges: the performance scores significantly oscillate based on the type of data on which the model was last trained. 

The data that the model was last trained on can be regarded as a fine-tuning step. Thus, we make the following observations:

\textbf{The base model achieves better performance on BLiMP Supplement during image-caption epochs.} As shown in subfigure \ref{training_dyn:sub1}, our base model obtains higher scores on BLiMP Supplement, a text-only benchmark evaluating grammar, at checkpoints saved during image-caption epochs compared to text-only epochs. We investigate the breakdown of the BLiMP Supplement scores and notice that the score difference for different epoch types stems from two subtasks, \textit{subject-auxiliary inversion} and \textit{turn-taking}. For these subtasks, the performance of our base model fluctuates by even $\sim$10\% between checkpoints. 
We thus investigate the log probability scores of our base model for each subtask example at checkpoints 500,000 (text-only epoch) and 550,000 (image-caption epoch), for which the former model is incorrect and the latter is correct. These two checkpoints present the highest difference in BLiMP Supplement scores (7.54\%). We make the following observations:
\begin{enumerate}
    \item For the \textit{subject-auxiliary inversion}, 68.2\% of the examples for which the model at checkpoint 500,000 (text-only epoch) is incorrect and the model at checkpoint 550,000 (image-caption epoch) is correct have the correct sentence of the pair starting with ``Is" followed by a noun phrase. For example, pairs such as (``Is the host expecting an award-winning director that hasn't finished dressing yet?", ``Hasn't the host is expecting an award-winning director that finished dressing yet?"). This contrasts with the distribution of the task, where 31.1\% of the pairs have the correct sentence starting with ``Is" followed by a noun phrase. We theorise that the Localized Narratives dataset supports the model at checkpoint 550,000 (image-caption epoch) in choosing the "Is" followed by noun phrase sentences with higher probability, which happen to be the correct sentences in these pairs. That is because there are 706,251 constructions of the form ``there is" followed by a noun phrase in the Localized Narratives dataset. We hypothesise that as a result, our model learns that the pattern ``is" followed by a noun phrase is more likely during the image-caption epochs.

    \item For the \textit{turn taking} subtask, even if the model at checkpoint 550,000 (image-caption epoch) chooses the correct sentence more often, it does so with little confidence. For most examples for which checkpoint 550,000 (image-caption epoch) is correct and checkpoint 500,000 (text-only epoch) is not, the log probability difference between the correct and incorrect sentence of the former checkpoint is less than 2 points. To put this in context, the log probability scores range between -89 and -156, for which 2 points represent 0.013\% to 0.0225\%. There is no noticeable pattern in the training data that can motivate the model's better performance during image-caption epochs on the the \textit{turn taking} subtask. We conclude that this behaviour requires further investigation which we leave for future work.

\end{enumerate}
\textbf{The base model achieves better performance on VQA during text-only epochs.} In figure \ref{training_dyn:sub2}, it can be noticed the score of our base model on VQA oscillates by 5\% to 10\% between text-only epochs and image-caption epochs. We theorise that the cause of these variations is the difference in textual data between the two types of epochs. There are no turn-taking constructions in the image-caption datasets, and the number of questions (25,300 question marks) is significantly lower than in the text-only datasets (1,083,559 question marks). However, both are present in the format of the VQA text data. Therefore, we conclude that the image-caption datasets support the VQA task less due to differences in the text format. We argue that for a high score on VQA during image-caption epochs, the image-caption datasets should contain samples similar to the task.

\textbf{The alternation between text-only and image-caption epochs has little to no effect on the BLiMP, EWoK and Winoground benchmarks for the base model.} As shown in figure \ref{fig:training_dyn}, there is little oscillation between text-only and image-caption epochs on the BLiMP benchmark, suggesting that the text-only dataset supports the model better for this task, but the score generally increases. There are no noticeable patterns for EWoK or Winoground.

\textbf{Note:} The reason the scores in all benchmarks stabilise after checkpoint 800,000 is because of the small learning rate (5e-5) combined with the learning rate schedule (cosine annealing) we chose for training. After checkpoint 800,000, the learning rate gradually decreases from 1e-5 to 0, which has little effect on the gradients.

\section{Correlation to Imageability Scores} \label{appendix:imageability}

\begin{table}[h]
\centering
\label{tab:gate_by_bins}
\begin{tabular}{lll}
\toprule
 \textbf{Category} & \multicolumn{2}{l}{\textbf{Gate Selection}}  \\

 & Mean (SD) & \# words \\
\midrule

Very Low ($<$342) & 0.427 (0.143) & 401  \\
Low (342-450) & 0.481 (0.130) & 96 \\
High (450-558) & 0.378 (0.126) & 78 \\
Very High ($>$558) & 0.351 (0.155) & 140  \\

\bottomrule
\multicolumn{3}{l}{\footnotesize Categories defined as $\mu \pm \sigma$ based on the MRC database. \vspace{-5pt}} \\
\multicolumn{3}{l}{\vspace{-8pt}\footnotesize SD = standard deviation.} 
\end{tabular}
\caption{Mean gate value per imageability bin for our base model (incorporating a soft gate per feature).}
\label{tab:imag_scores}
\end{table}

Table \ref{tab:imag_scores} summarises the mean gate values for our base model corresponding to meaningful imageability bins, defined using cutpoints at mean ± 1 SD from the MRC database \citep{coltheart1981mrc}.

\section{Experiments Setup} \label{experiments_setup}
For all architectural features and training strategies we define in subsections \ref{dynamic_gating}-\ref{data_curriculum} and \ref{ch4:dynamic_gating}-\ref{ch4:objective_functions}, we conduct experiments in the form of ablation studies in order to evaluate each potential improvement in isolation. We select our base architecture, described in section \ref{base_architecture}, and define one experiment per feature. We train each enhanced model in the same conditions and evaluate it on five BabyLM Challenge benchmarks: BLiMP, BLiMP Supplement, EwoK, Winoground and VQA.

\subsection{Base Model Implementation Details} \label{base_implementation_detals}

We implement our dual stream transformer in PyTorch \citep{imambi2021pytorch}, following the architecture we introduce in section \ref{base_architecture}. We summarise our hyperparameter choices for the base model in table \ref{tab:hyperparameters}.

We use pre-layer normalisation rather than post-layer normalisation in our implementation as previous research shows that pre-layer normalisation provides better training stability for networks larger than six layers \citep{takase2022b2t}, which is crucial given our limited training budget and inability to perform extensive hyperparameter searches. 

Following standard transformer design, we use residual connections around each sub-layer (feed-forward networks, self-attention and cross-attention).

While the image encoder may be over-parameterised for single token processing, empirical results validate this choice (Appendix \ref{appendix:diff_encoders}), and it ensures architectural consistency and directly comparable results for future extensions to patch-based visual inputs.

\subsection{Training Details} \label{training_implementation_details}

We train all of our models using the hyperparameters summarised in table \ref{tab:hyperparameters_training} using the BabyLM Challenge 2024 evaluation pipeline, with the exception of a few changes for the auxiliary objective function and data curriculum experiments. In the case of the auxiliary objective function experiments, we increase the batch size from 64 to 128 as a larger batch size is recommended for contrastive learning \citep{chen2020simple}, which results in a total of 553,510 steps. Due to computational constraints, we were not able to select a larger batch size. In the case of the data curriculum experiments, the data order differs according to the strategy we define for that experiment. For the model trained using LexiContrastive Grounding as the auxiliary function, we use weight tying as recommended in the original work \citep{zhuang2024lexicon}.

We select a learning rate of 5e-5 to ensure training stability, despite this being conservative for the model size. While alternating between text-only and image-caption epochs improves performance on the benchmarks (as shown in Appendix \ref{data_curriculum}), this training regime can cause gradient instability when transitioning between epoch types. Therefore, we adopt a lower learning rate to mitigate this risk.
\subsection{Data Pipeline Details}
The text-only training dataset is provided in \textit{.txt} files, while the multimodal one is provided in \textit{.json} files for the captions and \textit{.npy} for the image embeddings, where each row in the numpy array embeds one image as a global token of dimension 768. We load the text-only data as one training sample per line, and the image-caption data as one (image, caption) pair representing one training sample. We do not perform any preprocessing on either data.

For the text-only data and the captions, we tokenise the text using the GPT-2 tokeniser \citep{radford2019language}, as our model is autoregressive. We also add the \textit{BOS} and \textit{EOS} special tokens at the beginning and end of each text-only/caption sample, respectively.

We split the data into 80\% training, 10\% validation and 10\% held-out test sets. In order to ensure that all models are trained on the same data, we save the data split indices and reuse them for all experiments. We shuffle the training dataset independently for each run while maintaining consistent train/validation/test partitions.

\section{Design Choices for the Image Processing Pipeline} \label{appendix:diff_encoders}

Despite using only global image embeddings, we chose to implement an image encoder in our framework for the following reasons:
\begin{itemize}
    \item Future compatibility: We aim to develop future iterations of this framework that address current limitations by using patch tokens instead of global image embeddings. For comparable results, we choose to use an encoder for the CLS token as well, which benefits from feed-forward and normalisation layers, but not self-attention. The image encoder outputs a non-linear adaptation of pre-trained visual features and improves alignment with the text stream.
    
    \item Empirical performance:  We experimented with three variants: (1) directly using linearly projected DINOv2 embeddings, (2) applying a 2-layer multi-layer-perceptron (MLP), and (3) using the transformer encoder. The encoder variant demonstrated superior performance across benchmarks, which can be attributed to the encoder's deeper transformation capacity. The benchmark scores for the three variants are available in table \ref{tab:encoders}.

    \item Computational efficiency: An alternative to the image encoder is to import and fully or partially unfreeze the external pre-trained image encoder used in the BabyLM Challenge, \textit{facebook/dino2-base}\footnote{\url{https://huggingface.co/facebook/dinov2-base}}. However, this would require processing the raw images through the entire encoder (86.6 million parameters) during training, which would significantly increase the computational costs for data loading, forward passes (and backward passes if unfrozen) and memory usage. This approach contradicts the constraints of the challenge, which advocates for the fair use of computational resources. In contrast, a customisable image encoder component taking as input pre-computed embeddings can be modified based on the user's computational constraints.
\end{itemize}

Table \ref{tab:encoders} summarises the performance of the base model with different image processing pipelines, evaluated every 200,000 steps on BLiMP, BLiMP Supplement, EWoK, Winoground and VQA using the BabyLM Challenge 2024 evaluation pipeline. As shown, the results on BLiMP, BLiMP Supplement and VQA validate the use of a transformer encoder, for which the base model achieves the best scores. However, since a single image embedding cannot benefit from the self-attention mechanism, an MLP encoder suffices if computational resources are a constraint, achieving competitive performance .Besides the superior performance, the motivation for using a transformer encoder in this work was to enable a direct performance comparison with future iterations of the framework using patch-token embeddings.


\begin{table*}[h]
\centering
\begin{tabular}{ll}
\hline
\textbf{Model Hyperparameter} & \textbf{Value} \\
\hline
\multicolumn{2}{l}{\textit{Model Dimensions}} \\
Model dimension ($d_{model}$) & 768 \\
Hidden dimension & 3072 \\
Number of attention heads & 8 \\
Image encoder layers & 5 \\
Decoder layers & 8 \\
\hline
\multicolumn{2}{l}{\textit{Vocabulary \& Sequence}} \\
Vocabulary size & 50,260 (GPT-2 tokeniser \citep{radford2019language}) \\
Maximum sequence length & 128 \\
Special tokens & [PAD], [BOS], [EOS] \\
\hline
\multicolumn{2}{l}{\textit{Activation \& Regularisation}} \\
Activation function & GELU \citep{hendrycks2016gaussian} \\
Dropout rate & 0.1 \\
Layer normalisation & Pre-layer norm \\
Layer norm epsilon & 1e-5 (PyTorch default) \\
\hline
\multicolumn{2}{l}{\textit{Input Dimensions}} \\
DINOv2 embedding dimension & 768 \\
DINOv2 representation & \textit{CLS} token only \\
\hline
\multicolumn{2}{l}{\textit{Model Statistics}} \\
Total parameters & $\sim$198.5M \\
\hline
\end{tabular}
\caption{The hyperparameters list for our dual stream transformer base model.}
\label{tab:hyperparameters}
\end{table*}

\begin{table*}
\centering
\begin{tabular}{ll}
\hline
\textbf{Training Hyperparameter} & \textbf{Value} \\
\hline
Data order & Alternating between text-only and image-caption epochs \\
Number of epochs & 10 text-only and 10 image-caption \\
Total number of steps & 1,107,020 \\
Checkpoints saved & Every 50,000 steps \\
Batch size & 64 \\
Learning rate & 5e-5 \\
Learning rate schedule & Cosine annealing \\
Optimiser & AdamW with   $\beta_1=0.9, \beta_2=0.999, \epsilon=1e-8$ \\
Number of steps for warmup & $\sim$1\% \\
Weight decay & 0.01 \\
Gradient clipping norm & 1.0 \\
Main loss function & Cross-entropy \\
Random seed & 42 \\

\end{tabular}
\caption{The hyperparameters list for our base training regime.}
\label{tab:hyperparameters_training}
\end{table*}

\begin{table*}[h]
\centering
\begin{tabular}{@{}rlll@{}}
\hline
\# & Model Architecture & Model Hyperparams & Training Config \\
\hline
1 & Base model (§\ref{base_architecture}) & Default\textsuperscript{a} & Default\textsuperscript{b} \\
\hline
\multicolumn{4}{@{}l}{\textbf{Dynamic Gating}} \\
2 & Base model + no gate & Default\textsuperscript{a} & Default\textsuperscript{b} \\
3 & Base model + soft gate per feature & Default\textsuperscript{a} & Default\textsuperscript{b} \\
4 & Base model + soft gate per token & Default\textsuperscript{a} & Default\textsuperscript{b} \\
5 & Base model + hard gate per feature & Default\textsuperscript{a} & Default\textsuperscript{b} \\
6 & Base model + hard gate per token & Default\textsuperscript{a} & Default\textsuperscript{b} \\
\hline
\multicolumn{4}{@{}l}{\textbf{Feature Representation}} \\
7 & Base model + FiLM on text & Default\textsuperscript{a} & Default\textsuperscript{b} \\
8 & Base model + FiLM on image & Default\textsuperscript{a} & Default\textsuperscript{b} \\
9 & Base model + FiLM on cross-attention & Default\textsuperscript{a} & Default\textsuperscript{b} \\
10 & Base model + DyIntra on text & Default\textsuperscript{a} & Default\textsuperscript{b} \\
11 & Base model + DyIntra on image & Default\textsuperscript{a} & Default\textsuperscript{b} \\
12 & Base model + DyIntra on cross-attention & Default\textsuperscript{a} & Default\textsuperscript{b} \\
13 & Base model + Channel Attention & Default\textsuperscript{a} & Default\textsuperscript{b} \\
\hline
\multicolumn{4}{@{}l}{\textbf{Auxiliary Objectives}} \\
14 & Base model & Default\textsuperscript{a} & Default\textsuperscript{b} + CLIP (BS=128) \\
15 & Base model & Default\textsuperscript{a} + weight tying & Default\textsuperscript{b} + LCG (BS=128) \\ 
\hline
\multicolumn{4}{@{}l}{\textbf{Data Curriculum}} \\
16 & Base model & Default\textsuperscript{a} & Text-only → image-caption\textsuperscript{c} \\
17 & Base model & Default\textsuperscript{a} & Image-caption → text-only\textsuperscript{d} \\
18 & Base model & Default\textsuperscript{a} & Non-uniform mix\textsuperscript{e} \\
19 & Base model & Default\textsuperscript{a} & Uniform mix \\
\hline
\multicolumn{4}{@{}l}{\footnotesize \textsuperscript{a}As in Table~\ref{tab:hyperparameters} \quad \textsuperscript{b}As in Table~\ref{tab:hyperparameters_training} \quad BS = batch size} \\

\multicolumn{4}{@{}l}{\footnotesize \textsuperscript{c}First 10 epochs text-only, next 10 epochs image-caption} \\
\multicolumn{4}{@{}l}{\footnotesize \textsuperscript{d}First 10 epochs image-caption, next 10 epochs text-only} \\
\multicolumn{4}{@{}l}{\footnotesize \textsuperscript{e}Image-caption and text-only data non-uniformly mixed in same batch} \\
\end{tabular}
\caption{Summary of all the experiments we conduct in this work.}
\label{tab:experiment_summary}
\end{table*}

\begin{table*}[h]
\centering
\begin{tabular}{p{2cm}|c|c|c|c|c|c}
\hline
\textbf{Model} & \textbf{Checkpoint} & \textbf{BLiMP} &  \textbf{BLiMP S.} & \textbf{EWoK} & \textbf{Winoground} & \textbf{VQA} \\
\hline
\multirow{6}{*}{\shortstack{Base +\\No Encoder}} & 200K & 69.99 ± 0.17 & 54.03 ± 0.52 & 49.87 ± 0.57 & 51.74 ± 1.83 & 43.00 ± 0.31 \\
& 400K & 73.21 ± 0.16 & 53.33 ± 0.62 & 49.93 ± 0.57 & 52.68 ± 1.83 & 46.54 ± 0.31 \\
& 600K & 72.82 ± 0.16 & 52.89 ± 0.65 & 50.38 ± 0.57 & 52.95 ± 1.83 & 46.41 ± 0.31 \\
& 800K & 73.40 ± 0.16 & 53.69 ± 0.64 & 50.16 ± 0.57 & 52.41 ± 1.83 & 46.52 ± 0.31 \\
& 1M & 73.17 ± 0.16 & 53.62 ± 0.64 & 50.43 ± 0.57 & 52.14 ± 1.83 & 46.65 ± 0.31 \\
& 1.107M & 74.29 ± 0.16 & 55.63 ± 0.58 & 50.18 ± 0.57 & 51.21 ± 1.83 & 45.02 ± 0.31 \\
\hline
\multirow{6}{*}{\shortstack{Base +\\MLP\\Encoder}} & 200K & 70.66 ± 0.17 & 57.42 ± 0.49 & 50.03 ± 0.57 & 52.01 ± 1.83 & 41.72 ± 0.31 \\
& 400K & 73.47 ± 0.16 & 51.89 ± 0.66 & 49.96 ± 0.57 & 50.67 ± 1.83 & 48.41 ± 0.31 \\
& 600K & 73.20 ± 0.16 & 52.49 ± 0.68 & 50.02 ± 0.57 & 52.01 ± 1.83 & 43.32 ± 0.31 \\
& 800K & 74.06 ± 0.16 & 51.71 ± 0.67 & 50.47 ± 0.57 & 52.28 ± 1.83 & 48.18 ± 0.31 \\
& 1M & 73.71 ± 0.16 & 50.56 ± 0.67 & 50.21 ± 0.57 & 52.55 ± 1.83 & 48.54 ± 0.31 \\
& 1.107M & 74.35 ± 0.16 & 55.38 ± 0.60 & 50.21 ± 0.57 & 50.27 ± 1.83 & 49.83 ± 0.31 \\
\hline
\multirow{6}{*}{\shortstack{Base +\\Transformer\\Encoder}}  & 200K & 69.97 ± 0.17 & 55.02 ± 0.54 & 49.83 ± 0.57 & 51.07 ± 1.83 & 34.11 ± 0.32 \\
& 400K & 73.22 ± 0.16 & 51.88 ± 0.66 & 50.13 ± 0.57 & 50.94 ± 1.83 & 47.82 ± 0.31 \\
& 600K & 73.47 ± 0.16 & 53.08 ± 0.62 & 50.74 ± 0.57 & 51.61 ± 1.83 & 48.36 ± 0.31 \\
& 800K & 74.18 ± 0.16 & 52.69 ± 0.62 & 50.69 ± 0.57 & 52.41 ± 1.83 & 51.2 ± 0.31 \\
& 1M & 74.34 ± 0.16 & 54.00 ± 0.61 & 50.82 ± 0.57 & 52.41 ± 1.83 & 51.2 ± 0.31 \\
& 1.107M & 75.53 ± 0.16 & 55.71 ± 0.57 & 50.41 ± 0.57 & 51.74 ± 1.83 & 50.02 ± 0.31 \\
\hline
\end{tabular}
\caption{The performance of the base model with different image processing pipelines, evaluated using the 2024 BabyLM Challenge evaluation pipeline. The models are evaluated every 200,000 steps on BLiMP, BLiMP Supplement, EWoK, Winoground and VQA.}
\label{tab:encoders}
\end{table*}

\end{document}